\documentclass{article}

\usepackage[preprint,nonatbib]{neurips_2020}

\usepackage[utf8]{inputenc} % allow utf-8 input
\usepackage[T1]{fontenc}    % use 8-bit T1 fonts
\usepackage[colorlinks=true,linkcolor=red,citecolor=blue]{hyperref}       % hyperlinks
\usepackage{url}            % simple URL typesetting
\usepackage{booktabs}       % professional-quality tables
\usepackage{amsfonts}       % blackboard math symbols
\usepackage{nicefrac}       % compact symbols for 1/2, etc.
\usepackage{microtype}      % microtypography
\usepackage{caption}
\usepackage{subcaption}
\usepackage{wrapfig}
\usepackage[ruled,vlined]{algorithm2e}

\usepackage{amsfonts}
\usepackage{amssymb}
\usepackage{url}
\usepackage{xcolor}
\usepackage{kky}

\newcommand{\pd}{p_d(\xb)}          % data density
\newcommand{\MMD}{\mathbb{M}}       % Maximum Mean Discrepancy
\newcommand{\KSD}{\mathbb{S}}       % Kernel Stein Discrepancy
\newcommand{\KL}{\text{KL}}       % Kernel Stein Discrepancy

\newcommand{\bmphi}{\bm{\phi}}
\newcommand{\dkdxy}[2]{\nabla_{\xb,\yb}k(\xb,\yb)}

\newcommand{\SGLD}{{\sf SGLD}\xspace}
\newcommand{\SVGD}{{\sf SVGD}\xspace}
\newcommand{\ASGLD}{{\sf A-SGLD}\xspace}
\newcommand{\ASVGD}{{\sf A-SVGD}\xspace}
\newcommand{\NCKSVGD}{{\sf NCK-SVGD}\xspace}

\newcommand{\krbf}{k_{\text{rbf}}}
\newcommand{\kimq}{k_{\text{imq}}}

\providecommand{\scalT}[2]{\ensuremath{\left\langle{#1},{#2}\right\rangle}}

\def\argmax{\operatornamewithlimits{arg\,max}}
\def\argmin{\operatornamewithlimits{arg\,min}}

\title{Kernel Stein Generative Modeling}

\author{
    Wei-Cheng Chang$^1$ \hspace{0.7em} Chun-Liang Li$^2$   \hspace{0.7em} Youssef Mroueh$^3$  \hspace{0.7em} Yiming Yang$^1$ \\
    $^1$ Carnegie Mellon University, \hspace{0.7em} $^2$ Google, \hspace{0.7em} $^3$ IBM Research \\
    \texttt{\{wchang2,yiming\}@cs.cmu.edu} \hspace{0.7em} \texttt{chunliang@google.com} \hspace{0.7em} \texttt{mroueh@us.ibm.com}
}

\begin{document}

\maketitle

\begin{abstract}
%%%% ICML version %%%%
% We are interested in a novel principle of diverse data generation based  sampling from gradients of the data distribution (score function).
% %
% Two building blocks are at the heart of this paradigm. The first is estimating score functions of data distributions. The second is a sampling procedure that transports samples iteratively towards high density regions while maintaining certain diversity among the samples. 
% %
% For score estimation, we propose Kernel Stein Denoising Score Matching (KS-DSM), a variant of DSM that relies on the Kernel Stein discrepancy as a means for comparing score functions. For sampling, we use annealed Kernel Stein variational gradient descent (SVGD). 
% %
% Our models produce samples comparable to GANs on standard computer vision benchmarks.
% % up to you MNIST, CIFAR-10, and CelebA datasets.
% More interestingly, we show that KS-DSM and SVGD with a different choice of kernels offer a flexible trade-off between image quality and sample diversity.

%%%% NeurIPS version %%%%
% Our focus is EGM
We are interested in gradient-based \textit{Explicit Generative Modeling} where samples can be derived from iterative gradient updates based on an estimate of the score function of the data distribution.
Recent advances in Stochastic Gradient Langevin Dynamics (SGLD) demonstrates impressive results with energy-based models on high-dimensional and complex data distributions.
Stein Variational Gradient Descent (SVGD) is a deterministic sampling algorithm that iteratively transports a set of particles to approximate a given distribution, based on functional gradient descent that decreases the  KL divergence.
SVGD has promising results on several Bayesian inference applications.
However, applying SVGD on high-dimensional problems is still under-explored.
The goal of this work is to study high dimensional inference with SVGD. 
We first identify key challenges in practical kernel SVGD inference in high-dimension.
We propose noise conditional kernel SVGD (NCK-SVGD), that works in tandem with the recently introduced Noise Conditional Score Network estimator. 
NCK is crucial for successful inference with SVGD in high dimension, as it adapts the kernel to the noise level of the score estimate. As we anneal the noise,  NCK-SVGD targets the real data distribution.
We then extend the annealed SVGD with an entropic regularization. We show that this offers a flexible control between sample quality and diversity, and verify it empirically by precision and recall evaluations.
The NCK-SVGD produces samples comparable to GANs and annealed SGLD on computer vision benchmarks, including MNIST and CIFAR-10.

\end{abstract}

% main paper
%%%%%%%%%%%%%%%%%%%%%%
%%%% Introduction %%%%
%%%%%%%%%%%%%%%%%%%%%%
\section{Introduction}

% introduce IGMs versus EGMs, where sampling is key factor for the latter. 
%At the heart of generative models is the ability to draw novel samples indistinguishable from the data distribution.
Drawing novel samples from the data distribution is at the heart of generative models.  %which can be 
Existing work can be put into two categories, namely the \textit{Implicit Generative Models} (IGM) and the \textit{Explicit Generative Models} (EGM).
%For example, \textit{Implicit Generative Models} (IGM) such as generative adversarial networks~\cite{goodfellow2014generative} (GAN) learn to transform a simple source distribution to the target data distribution by minimizing $f$-divergence~\cite{nowozin2016f} or integral probability metric~\cite{arjovsky2017wasserstein,li2017mmd} between model and data distribution.
The generative adversarial networks (GAN) ~\cite{goodfellow2014generative} are representative examples of IGM,
which learns to transform simple source distributions to target data distributions by minimizing $f$-divergence~\cite{nowozin2016f} or integral probability metrics~\cite{arjovsky2017wasserstein,li2017mmd} between the model and the data distribution.
On the other hand, EGMs typically optimize the likelihood of non-normalized density models (e.g., energy-based models~\cite{lecun2006tutorial}) or learn score functions (i.e., 
gradient of log-density) %w.r.t. the input data
~\cite{hyvarinen2005estimation,vincent2011connection}.
Because of the explicitly modeling of densities or gradient of log-densities, EGM is still favorable for a wide range of applications such as anomaly detection~\cite{du2019implicit,grathwohl2019your}, image processing~\cite{song2019generative} and more.
However, the generative capability of many EGMs is not as competitive with GAN on high-dimensional distributions, such as images.
This paper focuses on an in-depth study in the EGMs.
%Nonetheless, how to make EGM adequately sampling from high-dimensional space remains an open challenge.
%a crucial step that significantly impacts EGMs' performance.

% describe SGLD and anneal-SGLD, which is stochastic-based sampling
Recent advances in Stochastic Gradient Langevin Dynamics (SGLD)~\cite{welling2011bayesian} has led to certain success in EGMs, especially with energy-based models~\cite{du2019implicit,nijkamp2019learning,grathwohl2019your} and score-based models~\cite{song2019sliced} for high-dimension inference tasks such as image generation.
As a stochastic optimization procedure, SGLD moves samples along the gradient of log-density, with carefully controlled diminishing random noise, which converges to the true data distribution with a theoretical guarantee. 
% NCSN here
The recent noise-conditioned score network ~\cite{song2019generative} estimates the score functions 
using perturbed data distributions with varying degrees of Gaussian noises. 
For inference, they consider annealed SGLD to produce impressive samples that are comparable to state-of-the-art (SOTA) GAN-based models on CIFAR-10.

% describe SVGD, and carefully state the argument how original SVGD people promote it over SGLD.
Another interesting sampling technique is Kernel Stein variational gradient descent (SVGD)~\cite{liu2016stein,liu2017stein}, which iteratively produce samples via deterministic updates that optimally reduce the KL divergence between the model and the target distribution.
The particles (samples) in SVGD interact with each other, simultaneously moving towards a high-density region following the gradients, and also pushing each other away due to a repulsive force induced from the kernels.
These interesting properties of SVGD has made it promising in various challenging applications such as Bayesian optimization~\cite{gong2019quantile}, deep probabilistic models~\cite{wang2016learning,feng2017learning,pu2017vae}, and reinforcement learning~\cite{haarnoja2017reinforcement}.

% Our motivation for SVGD
%However, 
Despite the attractive nature of SVGD, how to make its inference effective and scalable for complex high-dimensional data distributions is an open question that have not been studied in sufficient depth. % has been less studied and remains under-explored.
One major challenge in high-dimensional inference is to deal with multi-modal distributions with many of low-density regions, where SVGD can fail even on simple Gaussian mixtures.
A remedy for this problem is to use ``noise-annealing''.
However, such a relatively simple solution may still lead to deteriorating performance along with the increased dimensionality of data.

In this paper, we aim to significantly enhance the capability of SVGD for sampling from complex and high-dimensional distributions.
Specifically, we propose a novel method, namely the Noise Conditional Kernel SVGD or \NCKSVGD in short, where the kernels are conditionally learned or selected based on the perturbed data distributions.
Our main contributions can be summarized in three folds.
Firstly, we propose to learn the parameterized kernels with noise-conditional auto-encoders, which captures shared visual properties of sampled data at different noise levels.
Secondly, we introduce \NCKSVGD with an additional entropy regularization, for flexible control of the trade-off between sample quality and diversity, which is quantitatively evaluated with precision and recall curves.
Thirdly, the proposed \NCKSVGD achieves a new SOTA FID score of $21.95$ on CIFAR-10 within the EGM family and is comparable to the results of GAN-based models in the IGM family.
%\yy{To our knowledge, it is the first time that a EGM model achieves the comparable performance scores as the best of IGM models, with the capability of explicit density estimation which IGM models cannot offer.}
Our work shows that high dimensional inference can be successfully achieved with SVGD.   

%%%%%%%%%%%%%%%%%%%%%%
%%%% Background.  %%%%
%%%%%%%%%%%%%%%%%%%%%%
\section{Background}
\label{sec:background}
In this section, we review Stein Variational Gradient Descent (SVGD) for drawing samples from target distributions
and describe how to estimate score functions via a recent advance in Noise Conditional Score Network (NCSN).

%\subsection{Stein Variational Gradient Descent (SVGD) }
\paragraph{Stein Variational Gradient Descent (SVGD)}
Let $\pd$ be a positive and continuously differentiable probability density function on $\RR^d$.
For simplicity, we denote $\pd$ as $p$ in the following derivation.
SVGD~\cite{liu2016stein,liu2017stein} aims to find a set of particles $\{\xb_i\}_{i=1}^n \in \RR^d$ to approximate $p$,
such that the empirical distribution $q(\xb) = \sum_i \delta(\xb-\xb_i)$ of the particles weakly converges to $p$ when $n$ is large.
Here $\delta(\cdot)$ denotes the Dirac delta function.

To achieve this, SVGD begins with a set of initial particles from $q$, and iteratively updates them with a deterministic transformation function $T_{\bmphi}(\cdot)$:
%(\peter{revise needed on continuous form,here keep discrete , only equation2  you need it in continuous, let us verify it })
\begin{equation}
    \xb' \leftarrow T_{\bmphi}(\xb)= \xb + \epsilon \bmphi(\xb),
    %\quad 
    \text{ let }
    \bmphi^*_{q,p}=\argmax_{\bmphi \in \Hcal}
    \Big\{ 
        -\frac{d}{d\epsilon} \KL \big( T_{\bmphi}(q) \ || \ p \big) \Big|_{\epsilon=0}
        \text{ s.t. } \| \bmphi \|_{\Hcal} \leq 1
    \Big\},
    \label{eq:svgd-KL}
\end{equation}
where $T_{\bmphi}(q)$ is the measure of the updated particles $\xb'=T_{\bmphi}(\xb)$ and  $\bmphi_{q,p}^*: \RR^d \rightarrow \RR^d$ is the optimal transformation function maximally decreasing the KL divergence between the target data distribution $p$ and the transformed particle distribution $T_{\bmphi}(q)$, for $\bmphi$ in the unit ball in the RKHS.

By letting $\epsilon$ go to zero, the continuous Stein descent is therefore given by $dX_{t}=\bmphi^*_{q_t,p}(X_t) dt$, where $q_{t}$ is the density of $X_t$. A key observation from~\cite{liu2017stein} is that, under some mild conditions, the negative gradient of KL divergence in Eq.~\eqref{eq:svgd-KL} is exactly equal to the square \textit{Kernel Stein Discrepancy} (KSD):
\begin{equation}
    -\frac{d}{dt} \KL(q_t \ || \ p) 
    = \KSD(q_t \ || \ p)^2
    = \max_{\bmphi \in \Fcal} \big( \EE_{\xb \sim q_t} [\Acal_p \bmphi(x)] \big)^2,
    \label{eq:svgd-ksd}
\end{equation}
where 
$\Acal_p \bmphi(\xb) = \nabla \log p(\xb)^\top \bmphi(\xb) + \nabla \cdot \bmphi(\xb)$,
$\nabla \cdot \bmphi = \sum_{j=1}^d \partial_{\xb_j} \bmphi_j(\xb)$ and 
$\Acal_p$ is the Stein operator that maps a vector-valued function $\bmphi$ to a scalar-valued function $\Acal_p\bmphi$.

KSD $\KSD(q \ || \ p)$ provides a discrepancy measure between $q$ and $p$ and $\KSD(q \ || \ p) = 0 \ \text{iff} \ q=p$ given that $\Hcal$ is sufficiently large.
By taking $\Hcal$ to be a reproducing kernel Hilbert space (RKHS), KSD provides a closed-form solution of Eq.~\eqref{eq:svgd-ksd}.
Specifically, let $\Hcal_0$ be a RKHS of scalar-valued functions with a positive definite kernel $k(\xb,\xb')$, and $\Hcal=\Hcal_0 \times \ldots, \Hcal_0$ the corresponding $d \times 1$ vector-valued RKHS.
The optimal solution of Eq.~\eqref{eq:svgd-ksd}~\cite{liu2016kernelized,chwialkowski2016kernel} is $\KSD(q \ || \ p) = \|\bmphi_{q,p}^*(\cdot)\|_{\Hcal}$ where  
\begin{equation}
    \bmphi_{q,p}^*(\cdot) \propto 
    \EE_{\xb \sim q} [\Acal_p k(\xb,\cdot)]
    = \EE_{\xb \sim q} [ \nabla \log p(\xb)^\top k(\xb,\cdot) + \nabla_{\xb}k(\xb,\cdot) ].    
\end{equation}

The remaining question is how to estimate the score function $s_p(\xb) = \nabla_{\xb} \log \pd$ based on the data $\xb\sim \pd$ without knowing $\pd$ for  generative modeling tasks.

%\subsection{ Score Estimation }
\paragraph{Score Estimation}
To circumvent the expensive Monte Carlo Markov Chain (MCMC) sampling or the intractable partition function when estimating the non-normalized probability density (e.g., energy-based models), 
%Score Matching~\cite{hyvarinen2005estimation} aims to explicitly match the model score function $s_{\theta}(\xb): \RR^d \rightarrow \RR^d$ to the data score function (i.e., $\frac{1}{2} \EE_{\pd}[ \|s_{\theta}(\xb) - \nabla_{\xb}\log \pd\|_2^2 ]$),
Score Matching~\cite{hyvarinen2005estimation} directly estimates the score by $\arg\min_\theta \frac{1}{2} \EE_{\pd}[ \|s_{\theta}(\xb) - \nabla_{\xb}\log \pd\|_2^2 ]$, where $s_{\theta}(\xb): \RR^d \rightarrow \RR^d$. 
To train the model $s_{\theta}(\xb)$, we use the equivalent objective
\begin{equation}
    \argmin_\theta\EE_{\pd} \Big[ \text{tr}\big( \nabla_{\xb} s_{\theta}(\xb) \big) + \frac{1}{2} \| s_{\theta}(\xb) \|_2^2 \Big]
\end{equation}
without the need of accessing $\nabla_{\xb}\log \pd$. 
However, score matching is not scalable to deep neural network and high-dimensional data~\cite{song2019sliced}
due to the expensive computation of $\text{tr}(\nabla_{\xb}s_{\theta}(\xb))$.

To overcome this issue, denoising score matching (DSM)~\cite{vincent2011connection} instead matches $s_{\theta}(\xb)$ to a non-parametric kernel density estimator (KDE) 
(i.e.,
$\frac{1}{2} \EE_{p_{\sigma}(\tilde{\xb})}
\left[
    \| s_{\theta}(\tilde{\xb}) - \nabla_{\tilde{\xb}} \log p_{\sigma}(\tilde{\xb}) \|^2
\right]
$)
where $p_{\sigma}(\tilde{\xb}) = \frac{1}{n}\sum_{i=1}^n p_{\sigma}(\tilde{\xb}|\xb_i)$
and $p_{\sigma}(\tilde{\xb}|\xb) \propto \exp(-\|\tilde{\xb}-\xb\|^2/2\sigma^2)$ is a smoothing kernel 
with isotropic Gaussian of variance $\sigma^2$.
\cite{vincent2011connection} further show that the objective is equivalent to the following
\begin{equation}
    \frac{1}{2} \EE_{p_{\sigma}(\tilde{\xb}|\xb)\pd}
        \left[ 
        \| s_{\theta}(\tilde{\xb}) - \nabla_{\tilde{\xb}} \log p_{\sigma}(\tilde{\xb}|\xb) \|^2
        \right], 
    \label{eq:dsm-v1}    
\end{equation}
where the target $\nabla_{\tilde{\xb}} \log p_{\sigma}(\tilde{\xb}|\xb) = (\xb-\tilde{\xb})/\sigma^2$ has a simple closed-form. 
We can interpret Eq.~\eqref{eq:dsm-v1} as learning a score function $s_{\theta}(\tilde{\xb})$ to move noisy input $\tilde{\xb}$ toward the clean data $\xb$.

One caveat of the DSM is that the optimal score $s_{\theta^*}(\xb) = \nabla_{\xb} \log p_{\sigma}(\xb) \approx \nabla_{\xb}\pd$ is true only when the noise $\sigma$ is small enough.
However, learning the score function with the single-noise perturbed data distribution will lead to inaccurate score estimation in the low data density region on high-dimension data space, which could be severe due to the low-dimensional manifold assumption.
Thus, \cite{song2019generative} propose learning a noise-conditional score network (NCSN) based on multiple perturbed data distributions with Gaussian noises of varying magnitudes:
\begin{equation}
    \frac{1}{2L}\sum_{l=1}^L \sigma_l^2 \cdot
    \EE_{p_{\sigma_l}(\tilde{\xb}|\xb)\pd}
        \left[ 
        \| s_{\theta}(\tilde{\xb};\sigma_l) - \nabla_{\tilde{\xb}} \log p_{\sigma_l}(\tilde{\xb}|\xb) \|^2
        \right], 
    \label{eq:ncsn}
\end{equation}
where $\{\sigma_l\}_{l=1}^L$ is a positive geometric sequence such that $\sigma_1$ is large enough to mitigate the low density region on high-dimension space and $\sigma_L$ is small enough to minimize the effect of perturbed data.
Note that $\sigma_l^2$ is to balance the scale of loss function for each noise level $\sigma_l$.

After learning the noise conditional score functions $s_{\theta}(\tilde{\xb},\sigma)$,  
\cite{song2019generative} conduct anneal SGLD to draw samples from a sequence of the model's score, under annealed noise levels $\sigma_1>\dots > \sigma_L$, which is
\begin{equation}
    \xb_{t+1} \leftarrow \xb_t + \frac{\eta_l}{2} s_{\theta}(\xb_t,\sigma_l) + \alpha \sqrt{\eta_l} \zb_t,
    \quad \forall t=1,\ldots,T \ \text{and} \ l=1,\ldots,L,
    \label{eq:sgld}
\end{equation}
where $\zb_t \sim \Ncal(0,1)$ is the standard normal noise and $\alpha$ is a constant controlling the diversity of SGLD.
It is crucial to choose the learning rate $\eta_l = \eta_0 \cdot \sigma_l^2/\sigma_L^2$,
which balances the scale between the gradient term $\|\eta_l s_{\theta}(\xb,\sigma_l)\|$ and the noise term $\|\sqrt{\eta_l}\zb_t\|$. See detailed explanation in ~\cite{song2019generative}.

%NCSN is able to generate high quality samples comparable to GANs on MNIST and CIFAR-10 datasets. Given the success of NCSN, we consider using the pre-trained score network $s_{\theta}(\xb, \sigma)$ for SVGD on the image generation tasks.

%%%%%%%%%%%%%%%%%%%%%%
%%%% Challenges.  %%%%
%%%%%%%%%%%%%%%%%%%%%%
\section{Challenges of SVGD in High-dimension}
\label{sec:challenges}
In this section, we provide a deeper analysis of SVGD on a toy mixture model with an imbalance mixture weights, as the dimension of the data distribution increases.

\begin{figure}
    \centering
    \begin{subfigure}{.325\textwidth}
        \centering
        \includegraphics[width=1.00\linewidth]{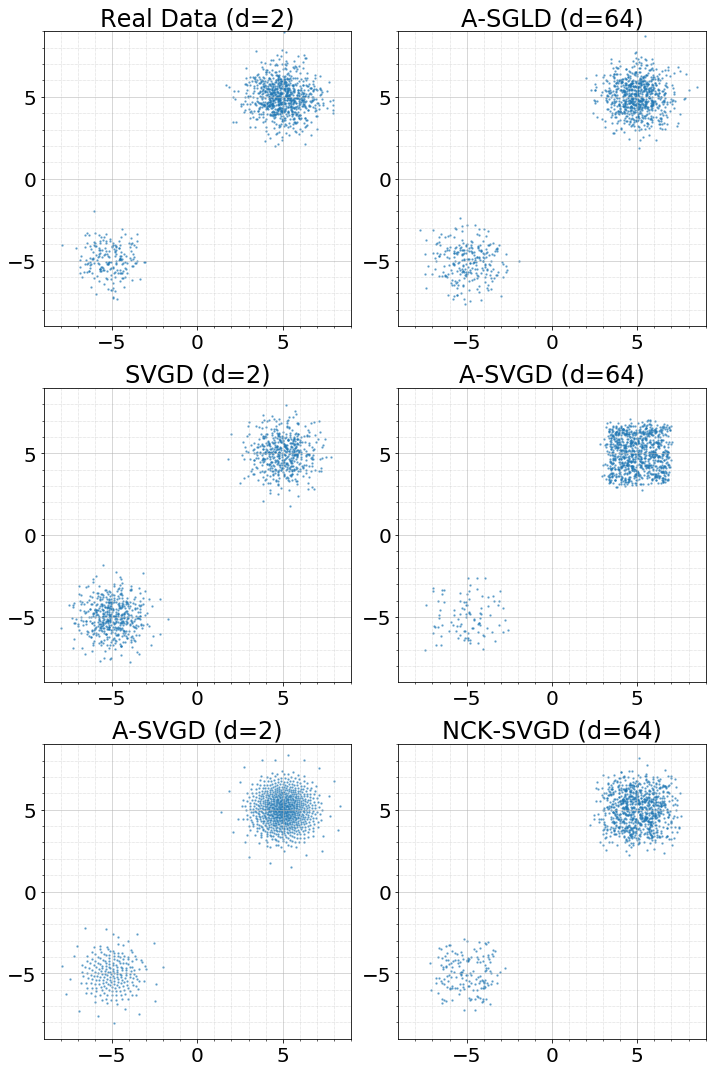}
        \caption{Samples visualization}
        \label{fig:toy-gmm-v0}
    \end{subfigure}
    \
    \begin{subfigure}{.475\textwidth}
        \centering
        \includegraphics[width=1.00\linewidth]{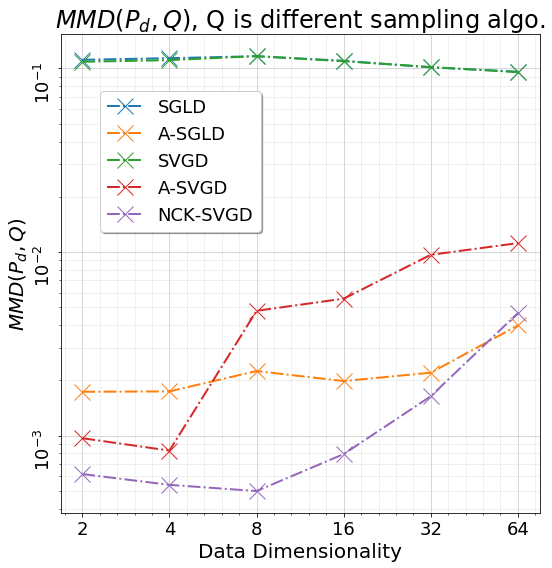}
        \caption{MMD loss versus data dimension}
        \label{fig:toy-gmm-v1}
    \end{subfigure}
    \vspace{-.25em}
    \caption{
        (a) Visualization of real data from $\pd$ and samples from different inference algorithms.
        The three figures in the left column are on 2-dimensional mixture of Gaussian.
        The right column is showing the results of different sampling algorithms on 64-dimensional distributions, and we visualize only the first two dimension, as the real-data distribution is isotropic.
        %Sub-figures at the second column are samples where inference algorithms applied to the 64-dimensional space, and we visualize only the first two dimension, as the real-data distribution is isotropic.
        (b) Maximum Mean Discrepancy (MMD) between the real-data samples ($P_d$) and generated samples ($Q$) from different inference algorithms,
        where \ASGLD means Anneal-SGLD, \ASVGD means Anneal-SVGD with a fixed kernel, and \NCKSVGD means Anneal-SVGD with noise-conditional kernels.
        Note that \SGLD (blue) and \SVGD (green) have alike samples with even mixture weights, so two curves overlapped.
    }
    \label{fig:toy-gmm}
    \vspace{-1em}
\end{figure}

\subsection{Mixture of Gaussian with Disjoint Support}
Consider a simple mixture distribution $\pd = \pi p_1(\xb) + (1-\pi)p_2(\xb)$, where $p_1(\xb)$ and $p_2(\xb)$ having disjoint, separated supports, and $\pi \in (0,1)$. 
\cite{li2019learning,song2019generative} demonstrate that score-based sampling methods such as \SGLD can not correctly recover the mixture weights of these two modes in reasonable time.
The reason is that the score function $\nabla_{\xb} \log \pd$ will be $\nabla_{\xb}\log p_1(\xb)$ in the support of $p_1$ and $\nabla_{\xb}\log p_2(\xb)$ in the support of $p_2$.
In either case, the score-based sampling algorithms are blind about the mixture weights, which may leads to samples with any reweighing of the components, depending on the initialization.
We now show that the Vanilla SVGD (\SVGD) also suffer from this issue on a 2-dimensional mixture of Gaussian $p_d(\xb) = 0.2\Ncal((-5,-5), I)+0.8\Ncal((5,5), I)$.
We use the ground truth scores (i.e., $\nabla_{\xb} \log \pd$) when sampling with \SVGD.
See the middle left of Figure~\ref{fig:toy-gmm-v0} for the samples and the green curve in Figure~\ref{fig:toy-gmm-v1} for the error (e.g., Maximum Mean Discrepancy) between the generated samples and the ground truth samples.
This phenomenon can also be explained by the objective of SVGD, as KL-divergence is not sensible to the mixture weights.
\begin{wrapfigure}{r}{4.5cm}
    \includegraphics[width=4.5cm]{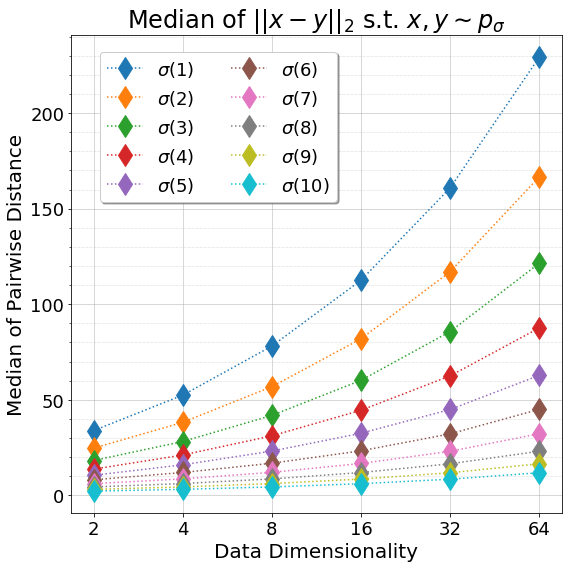}
    \caption{Medians of pairwise distance under different noise $\sigma(1)>\dots >\sigma(10)$ across $d$.}
    \label{fig:toy-gmm-v2}
    \vspace{-4em}
\end{wrapfigure}

\subsection{Anneal SVGD in High-dimension}
To overcome this issue, \cite{song2019generative} propose Anneal-SGLD (\ASGLD) with a sequence of noise-perturbed score functions, where injecting noises with varying magnitudes of the variance will enlarge the support of different mixture components, and mitigate the low-data density region in high-dimensional settings. 
We also perform Anneal-SVGD (\ASVGD) with a sequence of noise-perturbed score functions as used in \ASGLD, and consider the RBF kernel bandwidth to the median pairwise distance of samples from $\pd$.

For the low-dimensional case (e.g., $d=2$), \ASVGD produces samples that represents the correct mixture weights of $\pd$, as shown in bottom left of the Fig.~\ref{fig:toy-gmm-v0}. 
However, the performance of \ASVGD deteriorates as $d$ increases, as shown in the red curve of Fig.~\ref{fig:toy-gmm-v1}.
On the other hand, the performance of \ASGLD seems to be rather robust w.r.t. the increasing $d$ (i.e., the orange curve in Fig.~\ref{fig:toy-gmm-v1} and top-right samples in Fig.~\ref{fig:toy-gmm-v0}).

We argue that the failure of \ASVGD in high-dimension may be due to the inadequate kernel choice,
which are fixed to the median pairwise distance based on the real-data distribution $\pd$, regardless of the different noise level of $p_{\sigma}$.
In Fig.~\ref{fig:toy-gmm-v2}, we present $med(\{\|\xb-\yb\|_2\ \text{ s.t. } \xb,\yb \sim p_{\sigma} \})$, the median pairwise distance where samples are from different noise-perturbed $p_\sigma$, along with the increase of $d$.
We observe that, in low-dimensional setting (e.g., $d=2$), the medians of different $p_\sigma$ do not deviate too much. It explains the good performance of \ASVGD when $d<8$ in Fig.~\ref{fig:toy-gmm-v1}. 
Nevertheless, in  high-dimensional settings (e.g., $d=64$), the median differs a lot for the largest and smallest noise-perturbed data distribution (i.e., $p_{\sigma_1}$ v.s. $p_{\sigma_{10}}$).
The bandwidths suitable for large perturbed noise $\sigma_{10}$ no longer holds for small perturbed noise $\sigma_1$, hence limiting performance of fixed kernels. 

Following this insight, we propose noise-conditional kernels (NCK) for \ASVGD, namely \NCKSVGD, where the data-driven kernels $k(\xb,\xb;\sigma)$ is conditional on annealing noises $\sigma$ to dynamically adapt the varying scale changes. 
In this example, we can set the bandwidth of RBF kernel to be the median pairwise distance from noise-perturbed data distributions conditional on different noises $\sigma$.
See the samples visualization and performance of \NCKSVGD in Fig.~\ref{fig:toy-gmm-v0} and ~\ref{fig:toy-gmm-v1}, respectively.

%%%%%%%%%%%%%%%%%%%%%%
%%%% Framework.   %%%%
%%%%%%%%%%%%%%%%%%%%%%
\section{SVGD with Noise-conditional Kernels and Entropy Regularization}

In Fig.~\ref{fig:toy-gmm-v1}, we show a simple noise-conditional kernel (NCK) for \ASVGD leads to considerable gain on the Gaussian mixtures as described in Sec~\ref{sec:challenges}.
Here we discuss how to learn the NCK with deep neural networks for complex real-world data distributions.
What's more, we extend the proposed \NCKSVGD with entropy regularization for the diversity trade-off.

\subsection{Learning Deep Noise-conditional Kernels} 
Kernel selection, also known as kernel learning, is a critical ingredient in kernel methods, and has been actively studied in many applications such as generative modeling~\cite{sutherland2016generative,li2017mmd,binkowski2018demystifying,li2019implicit}, Bayesian non-parametric learning~\cite{wilson2016deep}, change-point detection~\cite{chang2019kernel}, statistical testing~\cite{li2019learning}, and more.

To make the kernel adapt to different noises,
we consider a deep NCK of the form
\begin{equation}
    k_{\psi}(\xb,\xb'; \sigma)
    = \krbf \big( E_{\psi}(\xb,\sigma), E_{\psi}(\xb',\sigma); \sigma \big)
    + \kimq \big( E_{\psi}(\xb,\sigma), E_{\psi}(\xb',\sigma); \sigma \big),
    \label{eq:deep-mok}
\end{equation}
where the Radius Basis (RBF) kernel $\krbf(\xb,\xb';\sigma)=\exp(-\gamma(\sigma) \|\xb-\xb'\|_2^2)$,
the inverse multiquadratic (IMQ) kernel $\kimq(\xb,\xb';\sigma)=(1.0 + \|\xb-\xb'\|_2^2)^{\tau(\sigma)}$,
and the learnable deep encoder $E_{\psi}(\xb,\sigma)$ with parameters $\psi$.
Note that the kernel hyper-parameters $\gamma$ and $\tau$ is also conditional on the noise $\sigma$.

Next, we learn the deep encoder $E_{\psi}(\xb,\sigma):\RR^d \rightarrow \RR^h$ via the noise-conditional auto-encoder framework to capture common visual semantic in noise-perturbed data distributions of $\{\sigma_l\}_{l=1}^L$:
\begin{equation}
    \frac{1}{2L}\sum_{l=1}^L \frac{1}{\sigma_l^2} \cdot
    \EE_{p_{\sigma_l}(\tilde{\xb}|\xb)\pd}
        \Big[ 
        \left\Vert
            D_{\varphi} \big( E_{\psi}(\tilde{\xb},\sigma_l), \sigma_l \big) -  \xb
        \right\Vert^2
        \Big], 
    \label{eq:nck-dae}
\end{equation}
where $D_{\varphi}(\xb,\sigma_l):\RR^h \rightarrow \RR^d$ is the corresponding noise-conditional decoder, and $h$ is the dimension of the code-space of auto-encoder.
Similar to the objective Eq.~\eqref{eq:ncsn}, the scaling constant $1/\sigma_l^2$ is to make the reconstruction loss of each noise level to be scale-balanced.

We note that the kernel learning via autoencoding is simple to train and is working well in our experimental study.
There are many recent advance in deep kernel learning~\cite{wilson2016deep,jacot2018neural,li2019learning}, which can potentially bring additional performance. We leave combining advanced kernel learning into the proposed algorithm as future work.

\subsection{Entropy Regularization and Diversity Trade-off}
Similarly to  \cite{wang2019nonlinear}, we propose to learn a distribution $q$ such that , for $\beta \geq 1$
$\min_{q}  \mathcal{F}_{\beta}(q)= \text{KL}(q,p) - (\beta-1) H(q)$.
The first term is the fidelity term the second term controls the diversity. 
We have 
%\vskip -0.20in
\begin{equation*}
\mathcal{F}_{\beta}(q)= \int q\log(q/p)+ (\beta-1) \int q \log(q)=\beta \int q\log(\nicefrac{q}{p^{\frac{1}{\beta}}})=\beta \text{KL}(q,p^{\frac{1}{\beta}}). 
\end{equation*}
\begin{remark}
Note that writing $\text{KL}(q,p^{\frac{1}{\beta}})$ is an abuse of notation, since $p^{\frac{1}{\beta}}$ is not normalized. Nevertheless as we will see next, sampling from $p^{\frac{1}{\beta}}$ can be seamlessly  achieved by an entropic regularization of the stein descent.  
\end{remark}
  
\begin{proposition}
Consider the continuous Stein descent 
$dX_t= \bmphi_{q_t,p,\beta}^*(X_t) dt $
where:
$\bmphi_{q_t,p,\beta}^*(x)= \mathbb{E}_{x'\sim q_t} s_{p}(x') K(x',x) + \beta \nabla_{x'}K(x', x) $
We have: 
$ \frac{d \mathcal{F}_{\beta}(q_t)}{dt} = - \beta^2 \mathbb{S}^2(q_{t},p^{\frac{1}{\beta}}). $
Hence the $\bmphi^*_{p,q_{t}\beta}$ is a descent direction for the entropy regularized KL divergence. More importantly the decreasing amount is the closeness in the Stein sense of $q_{t}$ to the smoothed distribution $p^{\frac{1}{\beta}}$.   
\end{proposition}

See Appendix~\ref{apx:entropy-reg-proof} for the detailed derivation.
From this proposition we see if $\beta$ is small, entropic regularized stein descent, will converge to $p^{\frac{1}{\beta}}$, and  will converge only to the high likelihood modes of the distribution and we have less diversity. If $\beta$ is high, on the other hand, we see that we will have more diversity but we will target a smoothed distribution $p^{\frac{1}{\beta}}$, since we have equivalently:
\begin{equation*}
 \frac{d \text{KL}(q,p^{\frac{1}{\beta}})}{dt}= -\beta \mathbb{S}^2(q_{t},p^{\frac{1}{\beta}}).
\end{equation*}
Hence $\beta$ controls the diversity of the sampling and the precision / recall of the sampling.
See Appendix~\ref{apx:entropy-reg-plot} for an illustration on a simple Gaussian mixtures and the MNIST dataset.

\begin{remark}
In Eq~\eqref{eq:sgld},
$\alpha$ plays the same role as $\beta$ and ensures convergence of SGLD to $p^{\frac{1}{\alpha}}$.
\end{remark}

\begin{algorithm}[H] 
	\caption{\NCKSVGD with Entropic Regularization.}
    \label{alg:MMD GAN}
 	
 	%\SetAlgoLined
    \SetKwInOut{Input}{input}\SetKwInOut{Output}{output}
    \Input{
        $\{\sigma_l\}_{l=1}^L$ the data-perturbing noises,
        $s_\theta(\xb,\sigma)$ the noise conditional score function,
        $k_{\psi}(\xb,\xb';\sigma)$ the noise conditional kernel,
        a set of initial particles $\{\xb_i^{(0)}\}_{i=1}^n$, \\
        the entropic regularizer $\beta$,
        an initial learning rate $\epsilon$, 
        and a maximum iteration $T$.}

 	\Output{A set of particles $\{\xb_i^{(T)}\}_{i=1}^n$ that approximates the target distribution.}
    
 	\For{$l\leftarrow 1$ \KwTo $L$}{
 	    $\eta_l \leftarrow \epsilon \cdot (\sigma_l/\sigma_L)^2$ \\
    	\For{$t\leftarrow 1$ \KwTo $T$}{
        	$\xb_i^{t} \leftarrow \xb_i^{(t-1)} + \eta_l \cdot \bmphi_{q_t,p,\beta}^*(\xb_i^{(t-1)})$,
        	\quad \text{where}
        	$\bmphi_{q_t,p,\beta}^*(\xb)=\frac{1}{n} \sum_{j=1}^n
        	\Big[
        	    k_\psi(\xb_j^{(t-1)}, \xb; \sigma_l)s_\theta(\xb,\sigma_l)
        	    + \beta \cdot \nabla_{\xb_j^{(t-1)}} k_\psi(\xb_j^{(t-1)}, \xb; \sigma_l)
        	\Big]$. \\
    	}
    	$\xb_i^{(0)} \leftarrow \xb_i^{(T)}, \forall i=1,\ldots, n$.
 	}
 	%\KwRet{$\{ \xb_i^{(T)} \}_{i=1}^n$}
\end{algorithm}

%%%%%%%%%%%%%%%%%%%%%%
%%%% Related Work. %%%
%%%%%%%%%%%%%%%%%%%%%%
\section{Related Work}
SVGD has been applied to deep generative models in various contexts~\cite{wang2016learning,feng2017learning,pu2017vae}.
Specifically, ~\cite{feng2017learning,pu2017vae} apply amortized SVGD to learn complex encoder functions in VAEs, which avoid the restricted assumption of the parametric prior such as Gaussian. 
\cite{wang2016learning} train stochastic neural samplers to mimic the SVGD dynamic and apply it for adversarial training the energy-based model with MLE objective, which avoids the expensive MCMC sampling.
Our proposed \NCKSVGD explicitly learn the noise-conditional score functions via matching annealed KDEs without any adversarial training.
For inference, \NCKSVGD leverage noise-conditional kernels to robustly interact with the noise-conditional score functions across different perturbing noise scales.
 
Recently,~\cite{wang2019nonlinear} present an entropy-regularized SVGD algorithm to learn diversified models, and apply it to the toy mixture of Gaussian and deep clustering.
The major difference between ~\cite{wang2019nonlinear} and this paper is two folds.
First, our analysis provides alternative insights on the entropy-regularized KL objective and show that it converges to $p^{1/\beta}$.
Second, the main application of \NCKSVGD is high-dimensional inference for image generation, which is more challenging than the applications presented in ~\cite{wang2019nonlinear}.

~\cite{ye2020stein} propose Stein self-repulsive dynamics, which integrates SGLD with SVGD to decrease the auto-correlation of Langevin Dynamics, hence potentially encourage diversified samples. 
Their work is complementary to our proposed \NCKSVGD, and is worth exploring as a future direction.

%%%%%%%%%%%%%%%%%%%%%%
%%%% Experiment.  %%%%
%%%%%%%%%%%%%%%%%%%%%%
\section{Experiments}
We begin with the experiment setup and show \NCKSVGD produces good quality images on MNIST/CIFAR-10 datasets as well as offers flexible control between the sample quality and diversity. 
\ASGLD is the primary competing method, and we use the recent state-of-the-art~\cite{song2019generative} for comparison. 

\textbf{Network Architecture}
For score network, we adapt noise-conditional score network (NCSN) ~\cite{song2019generative} for both \ASGLD and \NCKSVGD.
For the noise-conditional kernels, we use the same architecture as NCSN except for reducing the bottleneck layer width to a much smaller embedding size (e.g., from $32,768$ to $512$ on CIFAR-10).   
Please refer to Appendix~\ref{app:exp-setup-real} for more details.

\textbf{Kernel Design}
We consider a mixture of RBF and IMQ kernel on the data-space and code-space features,
as defined in Eq~\eqref{eq:deep-mok}.
The bandwidth of RBF kernel $\gamma(\sigma) = \gamma_0 / med(X_\sigma)$,
where $med(X_\sigma)$ denotes the median of samples' pairwise distance drawn from anneal data distributions $p_\sigma(\tilde{\xb}|\xb)$.

\textbf{Inference Hyper-parameters}
Following \cite{song2019generative}, we choose $L=10$ different noise levels where the standard deviations $\{\sigma_i\}_{i=1}^L$ is a geometric sequence with $\sigma_1=1$ and $\sigma_{10}=0.01$.
Note that Gaussian noise of $\sigma=0.01$ is almost indistinguishable to human eyes for image data.
%For \ASGLD, we choose $T=100$ and $\epsilon=2\times10^{-5}$ and $\alpha=\{0.1, 0.2, \ldots, 1.1, 1.2\}$.
For \NCKSVGD, we choose $n=128$, $T=50$, $\epsilon=\{2,4,6\}\times10^{-4}$, and $\beta=\{0.1, 0.2, \ldots, 2.9, 3.0\}$.

\textbf{Evaluation Metric}
We report the Inception~\cite{salimans2016improved} and FID~\cite{heusel2017gans} scores using $50$k samples.
In addition, we also present the Improved Precision Recall (IPR) curve~\cite{kynkaanniemi2019improved} to justify the impact of entropy regularization and kernel hyper-parameters on diversity versus quality trade-off.

\vspace{-0.1in}
\subsection{Quantitative Evaluation}

%%%% MNIST %%%%
\begin{figure*}
    \centering
    \begin{subfigure}{0.34\textwidth}
        \centering
        \includegraphics[width=1.00\linewidth]{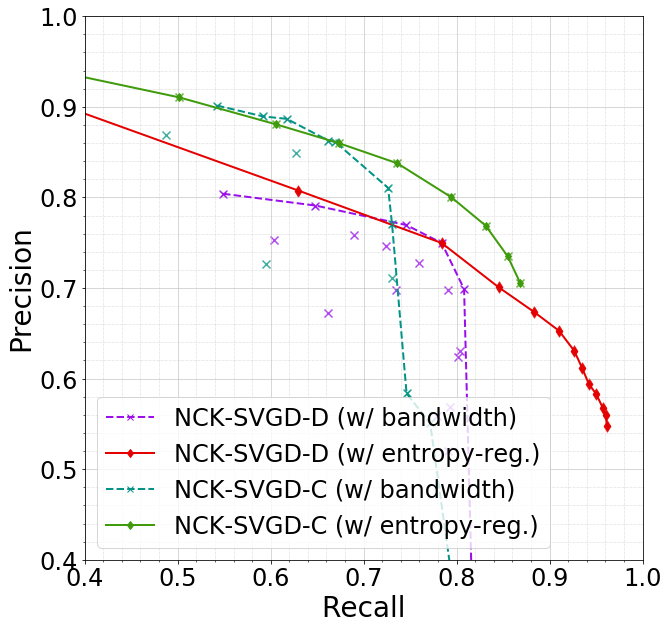}
        \caption{Ablation of \NCKSVGD}
        \label{fig:mnist-ipr-v0}
    \end{subfigure}
    \begin{subfigure}{0.34\textwidth}
        \centering
        \includegraphics[width=1.0\linewidth]{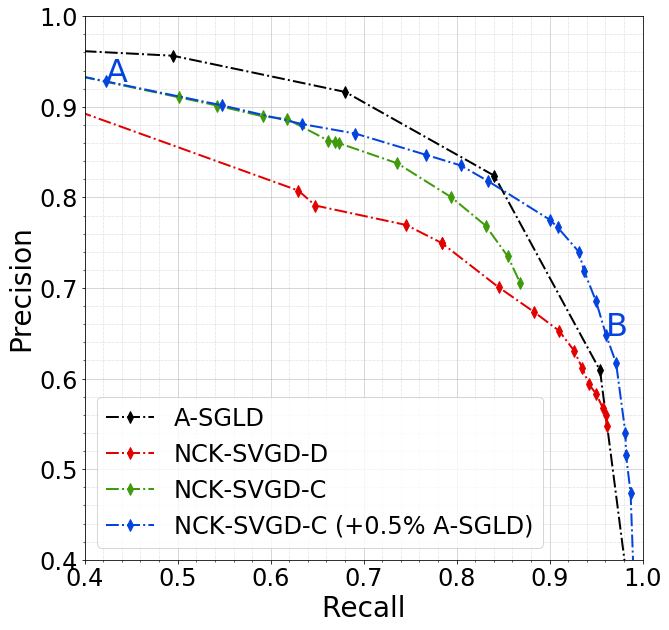}
        \caption{Comparison with \ASGLD}
        \label{fig:mnist-ipr-v1}
    \end{subfigure}
    \begin{subfigure}{0.3\textwidth}
        \centering
        \includegraphics[width=1.0\linewidth]{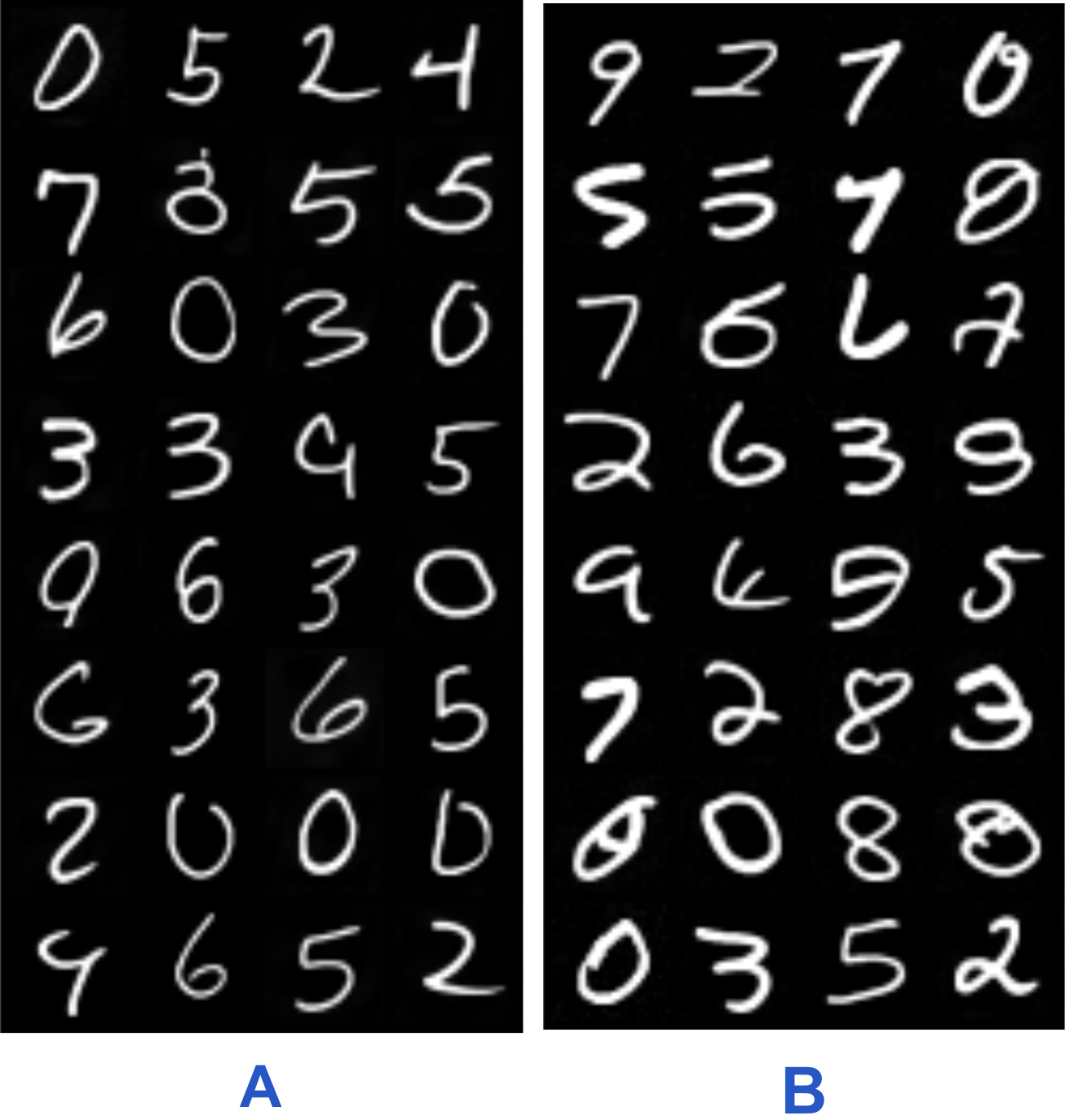}
        \caption{Representative samples}
        \label{fig:mnist-ipr-v2}
    \end{subfigure}
    \caption{
    MNIST experiment evaluated with Improved Precision and Recall (IPR).
    (a) \NCKSVGD-D (data-space kernels) and \NCKSVGD-C (code-space kernels) with varying kernel bandwidth ($\gamma$ from RBF kernel) and the entropy-regularizer $\alpha$. 
    (b) With $0.5\%$ of \ASGLD (i.e., 5 steps at noise-level $l=0$) as initialization, \NCKSVGD-C achieves higher recall than \ASGLD.
    (c) Uncurated samples of \NCKSVGD-C, including the high-precision/low-recall (A) to low-precision/high-recall (B). }
    \label{fig:mnist-exp}
    \vskip -0.15in
\end{figure*}
\paragraph{MNIST}
We analyze variants of \NCKSVGD quantitatively with IPR~\cite{kynkaanniemi2019improved} on MNIST, as shown in Fig.~\ref{fig:mnist-exp}.
\NCKSVGD-D denotes the \NCKSVGD with data-space kernels and \NCKSVGD-C as the \NCKSVGD with code-space kernels. 
We have three main observations.
First and foremost, both \NCKSVGD-D and \NCKSVGD-C demonstrate flexible control between the quality (i.e., Precision) and diversity (i.e., Recall) trade-off.
This finding aligns well with our theoretical analysis on the entropy regularization constant $\beta$ explicitly controls the samples diversity (i.e., the green and red curve in Fig.~\ref{fig:mnist-ipr-v0}).
Furthermore, the bandwidth $\gamma$ of RBF kernel also impacts the sample diversity, which attests the original study of SVGD on the repulsive term $\nabla_{\xb'}k(\xb',\xb)$~\cite{liu2017stein}.
Secondly, \NCKSVGD-C improves upon \NCKSVGD-D on the higher precision region, which justifies the advantages from using deep noise-conditional kernel with auto-encoder learning.  
Finally, when initializing with samples obtained from 5 out of 100 steps of \ASGLD at the first noise-level $\sigma_1$ (i.e., $0.5\%$ of total \ASGLD), \NCKSVGD-C achieves a higher recall compared to the SOTA \ASGLD.
%We will discuss the importance of initialization to the later section.

%%%% CIFAR-10 %%%%
\begin{figure}
    \centering
    \begin{minipage}{0.45\textwidth}
        \centering
        \captionsetup{type=table} %% tell latex to change to table
        \resizebox{0.93\columnwidth}{!}{%
        \begin{tabular}{ccc}
            \toprule
            Model                                       & Inception & FID       \\
            \midrule
            CIFAR-10 Unconditional \\
            \midrule        
            WGAN-GP~\cite{gulrajani2017improved}        & 7.86      & 36.4      \\
            MoLM~\cite{ravuri2018learning}              & 7.90      & 18.9      \\
            SN-GAN~\cite{miyato2018spectral}            & 8.22      & 21.7      \\
            ProgressGAN~\cite{karras2017progressive}    & 8.80      & -         \\
            \midrule
            %EBM (Single) ~\cite{du2019implicit}         & 6.02      & 40.58     \\
            EBM (Ensemble) ~\cite{du2019implicit}       & 6.78      & 38.2      \\
            Short-MCMC~\cite{nijkamp2019learning}       & 6.21      & -         \\
            \ASGLD~\cite{song2019generative}            & \textbf{8.87}      & 25.32     \\
            %\NCKSVGD-D                                  & 7.93      & 22.22     \\
            \NCKSVGD                                    & 8.20      & \textbf{21.95}     \\
            \midrule
            CIFAR-10 Conditional \\
            \midrule
            EBM~\cite{du2019implicit}                   & 8.30      & 37.9      \\
            JEM~\cite{grathwohl2019your}                & 8.76      & 38.4      \\
            SN-GAN~\cite{miyato2018spectral}            & 8.60      & 17.5      \\
            BigGAN~\cite{brock2018large}                & 9.22      & 14.73     \\
            \bottomrule
        \end{tabular}%
        }
        \caption{CIFAR-10 Inception and FID scores}
        \label{tab:cifar-is-fid}
    \end{minipage}
    \begin{minipage}{0.475\textwidth}
        \centering
        \includegraphics[width=0.92\linewidth]{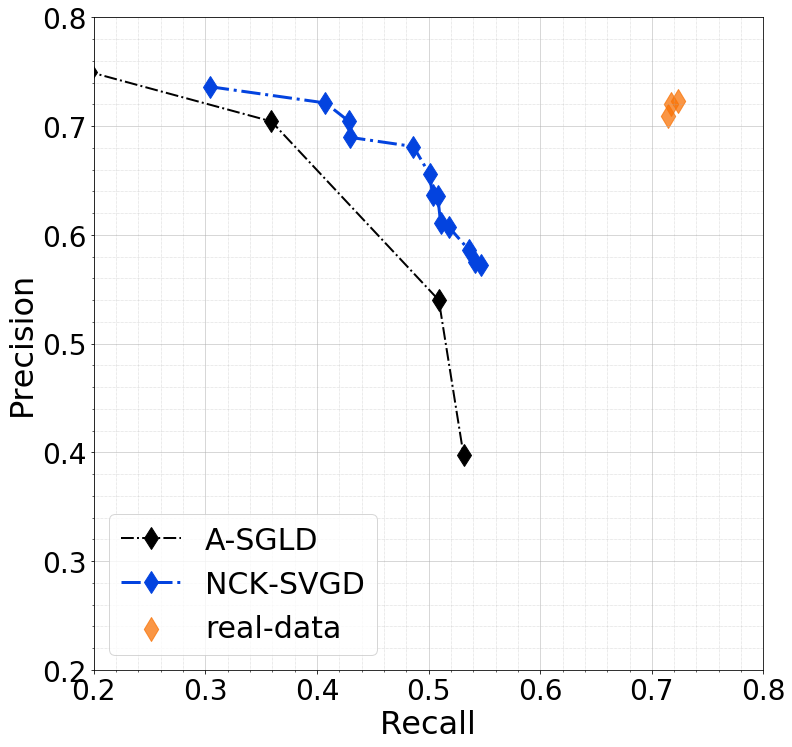}
        \caption{CIFAR-10 Precision-Recall Curve}\label{fig:cifar-ipr}
    \end{minipage}
\vskip -0.15in
\end{figure}

\textbf{CIFAR-10}
We compare the proposed \NCKSVGD (using code-space kernels) with two representative families of generative models: IGMs (e.g., WGAN-GP~\cite{gulrajani2017improved}, MoLM~\cite{ravuri2018learning}, SN-GAN~\cite{miyato2018spectral}, ProgressGAN~\cite{karras2017progressive}) and gradient-based EGMs (e.g., EBM~\cite{du2019implicit}, Short-run MCMC~\cite{nijkamp2019learning}, A-SGLD~\cite{song2019generative}, JEM~\cite{grathwohl2019your}).
See Tab.~\ref{tab:cifar-is-fid} for the Inception/FID scores and Fig.~\ref{fig:cifar-ipr} for the IPR curve.

Motivated from the MNIST experiment, we initialized the \NCKSVGD using \ASGLD samples generated from the first 5 noise-levels $\{\sigma_1, \ldots, \sigma_5\}$, then continue running \NCKSVGD for the remaining latter 5 noise-levels $\{\sigma_6, \ldots, \sigma_{10}\}$. This amounts to using $50\%$ of \ASGLD as initialization.

Comparing within the gradient-based EGMs (e.g., EBM, Short-run MCMC, and \ASGLD), \NCKSVGD achieves new SOTA FID score of $21.95$, which is considerably better than the competing \ASGLD and is even better than some \textit{class-conditional} EGMs.
The Inception score $8.20$ is also comparable to top existing methods, such as SN-GAN~\cite{miyato2018spectral}.

%It is argued that FID is a better evaluation metric for generative modeling and aligns well with coverage/recall.(\peter{any citation support this?})
From the IPR curve of Fig.~\ref{fig:cifar-ipr}, we see that \NCKSVGD improves over the \ASGLD with sizable margin, especially in the high recall region.
This again certifies the advantage of noise-conditional kernels and the entropy regularization indeed encourages samples with higher-recall.
\begin{figure*}
    \centering
    \begin{subfigure}{0.3\textwidth}
        \centering
        \includegraphics[width=1.00\linewidth]{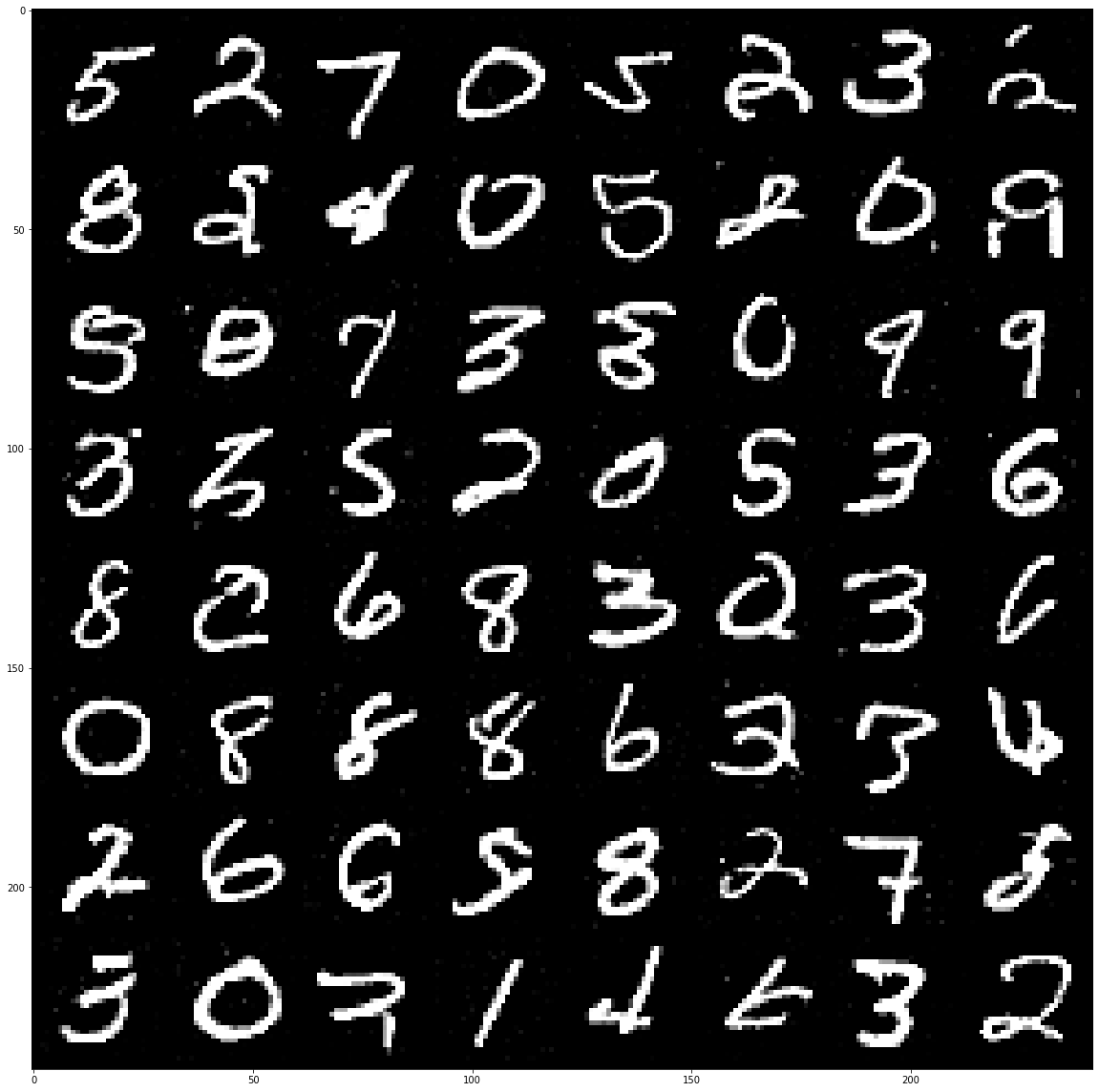}
        \caption{MNIST}
        \label{fig:mnist-samples-v0}
    \end{subfigure}
    \begin{subfigure}{0.3\textwidth}
        \centering
        \includegraphics[width=1.0\linewidth]{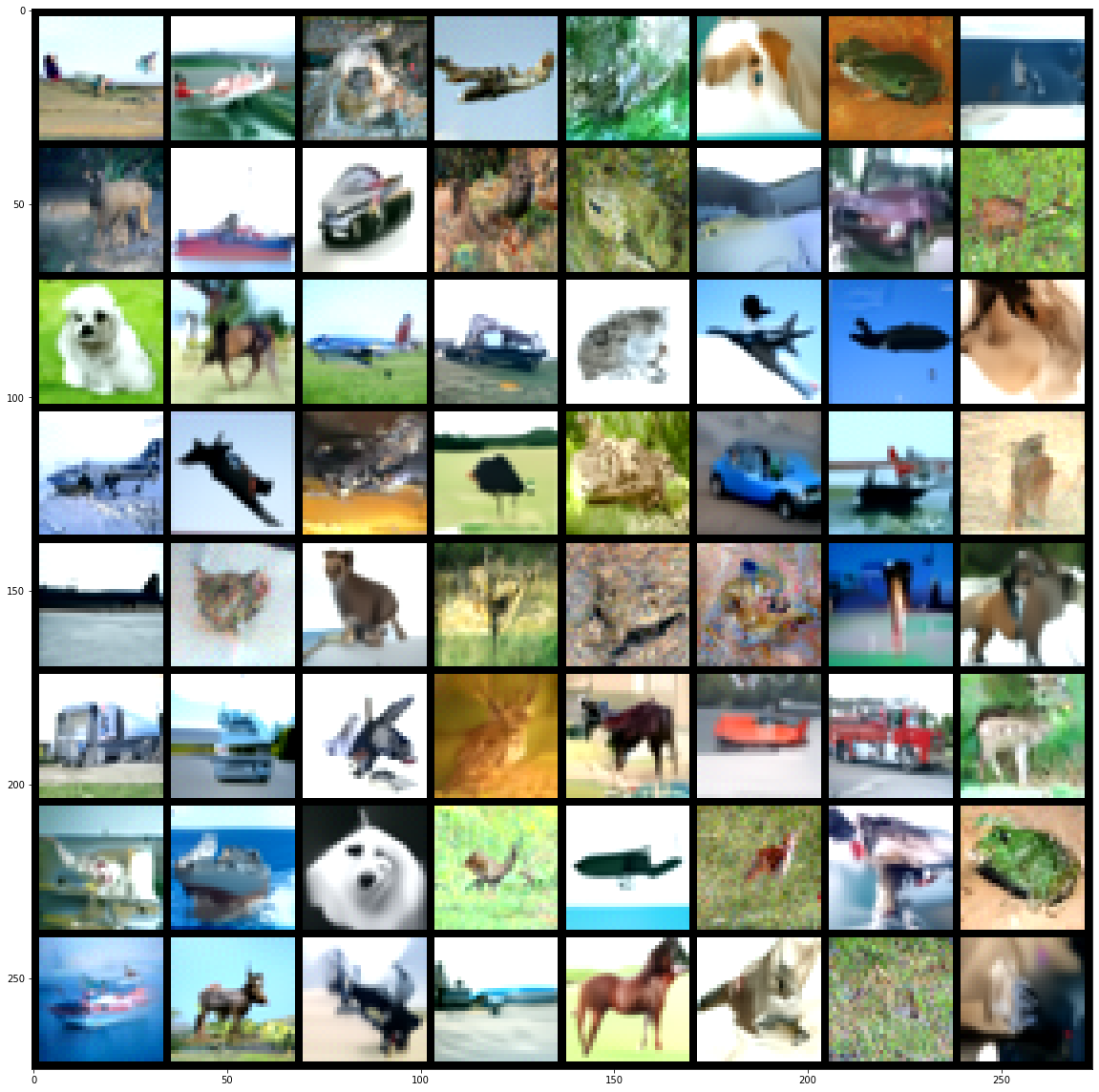}
        \caption{CIFAR-10}
        \label{fig:cifar10-samples-v1}
    \end{subfigure}
    \begin{subfigure}{0.35\textwidth}
        \centering
        \includegraphics[width=1.0\linewidth]{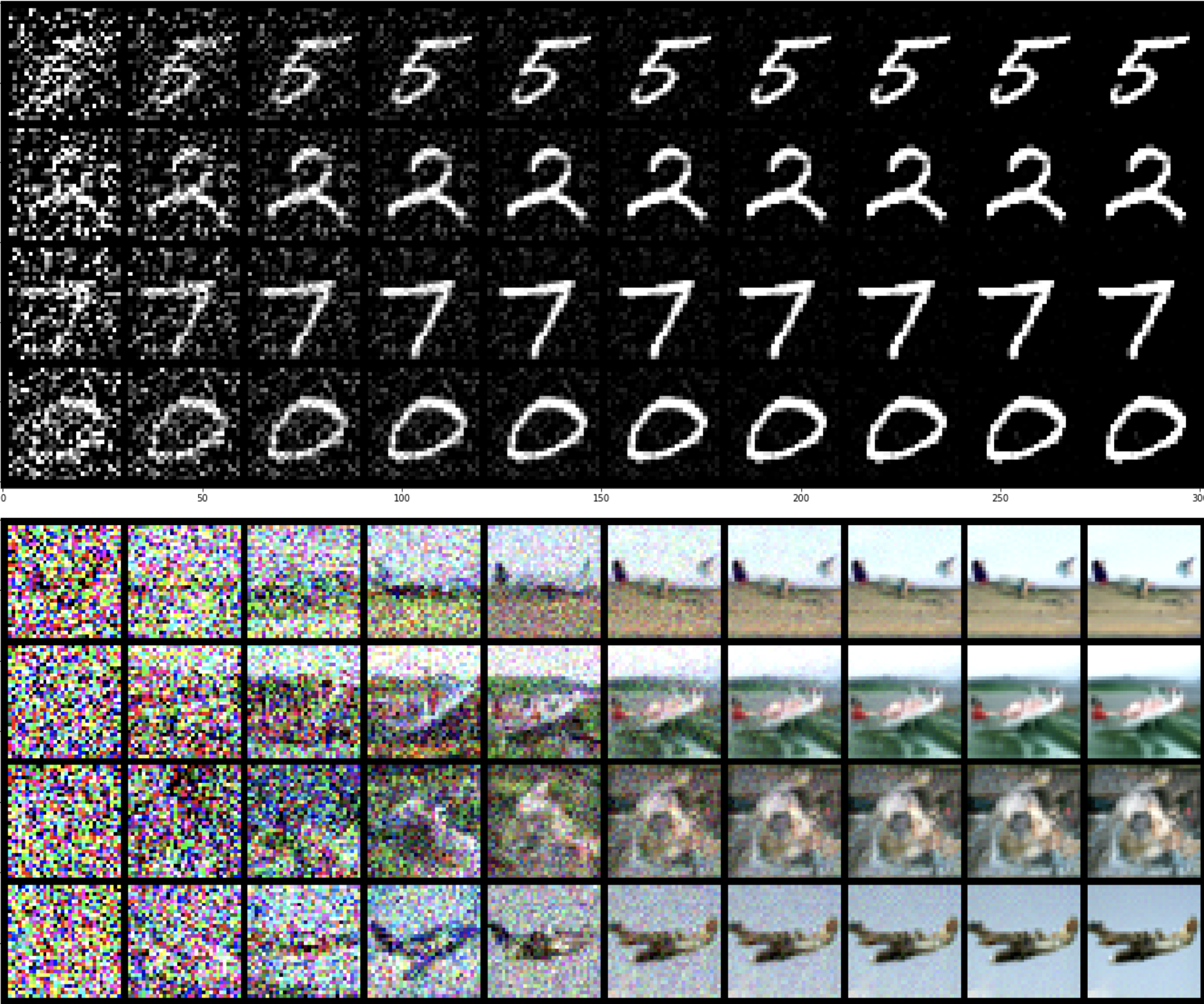}
        \caption{Intermediate samples}
        \label{fig:mnist-cifar10-gif}
    \end{subfigure}
    \caption{Uncurated samples generated from \NCKSVGD on MNIST and CIFAR-10 datasets.}
    \label{fig:mnist-cifar-samples}
\vskip -0.2in    
\end{figure*}

\subsection{Qualitative Analysis}
\vskip -0.1in 
We present the generated samples from \NCKSVGD in Fig.~\ref{fig:mnist-cifar-samples}.
Our generated images have higher or comparable quality to those from modern IGM like GANs or gradient-based EGMs.
To intuit the procedure of \NCKSVGD, we provide intermediate samples in Fig.~\ref{fig:mnist-cifar10-gif},
where each row shows how samples evolve from the initialized samples to high quality images.
We also compare \NCKSVGD against two SVGD baselines mentioned in Sec.~\ref{sec:challenges}, namely \SVGD and \ASGLD.
Due to the space limit, see the failure samples from the two baselines in Appendix~\ref{app:exp-baseline}.

%%%%%%%%%%%%%%%%%%%%%%
%%%% Conclusion.  %%%%
%%%%%%%%%%%%%%%%%%%%%%
\section{Conclusion}
\label{sec:conclusion}
In this paper, we presented \NCKSVGD, a diverse and deterministic sampling procedure of high-dimensional inference for image generations.
\NCKSVGD is competitive to advance stochastic MCMC methods such as \ASGLD, and reaching a lower FID scores of $21.95$.
In addition, \NCKSVGD with entropic regularization offers a flexible control between sample quality and diversity, which is quantitatively verified by the precision and recall curves.

\section*{Broader Impact}
%Soon by sampling from NCK-SVGD spaceX will reach diverse undiscovered planets.  
Recent development in generative models have begun to blur lines between machine and human generated content, creating an additional care to look at the ethical issues such as the copyright ownership of the AI generated art pieces, face-swapping of fake celebrity images for malicious usages, producing biased or offensive content reflective of the training data, and more.
Our \NCKSVGD framework, which performs annealed SVGD sampling via the score functions learned from data, is no exception.
Fortunately, one advantage of explicit generative model families including the proposed \NCKSVGD is having more explicit control over the iteratively sampling process, where we can examine if the intermediate samples violate any user specification or other ethical constraints.
On the other hand, implicit generative models such as GANs are less transparent in the generative process, which is a one-step evaluation of the complex generator network.

%%%%%%%%%%%%%%%%%%%
%%%% reference %%%%
%%%%%%%%%%%%%%%%%%%
\bibliography{sdp}
\bibliographystyle{unsrt}

% appendix
\appendix
\newpage
%\onecolumn

\section{Generative Modeling with Diversity Constraint}

\subsection{Technical Proof}
\label{apx:entropy-reg-proof}
We propose to learn a distribution $q$ such that , for $\beta \geq 1$
$$\min_{q}  \mathcal{F}_{\beta}(q)= \text{KL}(q,p) - (\beta-1) H(q)$$
The first term is the fidelity term the second term controls the diversity. 
We have 
$$\mathcal{F}_{\beta}(q)= \int q\log(q/p)+ (\beta-1) \int q \log(q)  $$ 
Its first variation is given by:
$$D_q \mathcal{F}(q)=  \nabla_x \log\left(\frac{q}{p}\right) + (\beta-1) \nabla_x \log(q)= \nabla_x \log \left(\frac{q^{\beta}}{p}\right)  $$
\begin{proposition}
Consider the continuous Stein descent 
$$dX_t= \bmphi^*_{p,q_t,\beta} (X_t) dt $$
where:
$$\bmphi^*_{p,q_t, \beta}(x)= \mathbb{E}_{x'\sim q_t} s_{p}(x') K(x',x) + \beta \nabla_{x'}K(x', x) $$
We have: 
\[ \frac{d \mathcal{F}_{\beta}(q_t)}{dt} = - \beta^2 \mathbb{S}^2(q_{t},p^{\frac{1}{\beta}}). \]
Hence the $f^*_{p,q_{t}\beta}$ is a descent direction for the entropy regularized KL divergence. More importantly the decreasing amount is the closeness in the Stein sense of $q_{t}$ to the smoothed distribution $p^{\frac{1}{\beta}}$.   
\end{proposition}

From this proposition we see if $\beta$ is small, entropic regularized stein descent, will converge to $p^{\frac{1}{\beta}}$, and  will converge only to the high likelihood modes of the distribution and we have less diversity. If $\beta$ is high, on the other hand, we see that we will have more diversity but we will target a smoothed distribution $p^{\frac{1}{\beta}}$. Hence $\beta$ controls the diversity of the sampling and the precision / recall of the sampling.

\begin{proof}
\noindent Let $q_{t}$ be the density of $q_{t}$, We have:
\begin{align*}
\frac{\partial q_t}{\partial t}&= -div(q_t \bmphi^*_{p,q_t,\beta} ) 
\end{align*}

It follows that we have: 
\begin{eqnarray*}
\frac{d \mathcal{F}_{\beta}(q_t)}{dt} &=& \int \scalT{ \nabla_x D_{q} \mathcal{F}(q_t)  }{\bmphi^*_{p,q_t,\beta} } q_t   \\
&=&  \int \scalT{ \nabla_x  \log(q_t^{\beta}/p)    }{\bmphi^*_{p,q_t,\beta} } q_t \\
&=&   \int \scalT{ \bmphi^*_{p,q_t,\beta} }{ \beta \nabla_x \log (q_t ) - s_p(x)}q_t\\
&=&  -   \left( \int  \scalT{s_p}{\bmphi^*_{p,q_t,\beta} }  q_t   - (\beta) \int \scalT{\nabla_x q_t}{f^*_{p,q_t, \beta}}
\right)\\
&=&   -    \left( \int  \scalT{s_p}{\bmphi^*_{p,q_t,\beta} }  q_t   + (\beta) \int  div(f^*_{p,q_t, \beta}) q_t\right ) \\
&& \text{(Using divergence theorem)}\\
&=&   -   \beta^2 \left( \int  \scalT{\frac{1}{\beta}s_p} {\frac{1}{\beta}\bmphi^*_{p,q_t,\beta} }  q_t   + \frac{1}{\beta} \int  div(f^*_{p,q_t, \beta}) q_t\right ) \\
&=&   - \beta^2 \mathbb{S}^2(q,p^{\frac{1}{\beta}})\\
&\leq& 0
\end{eqnarray*}

\end{proof}

\subsection{The Effect of Entropy Regularization on Precision/Recall}
\label{apx:entropy-reg-plot}

\begin{figure*}
    \centering
    \begin{subfigure}{0.51\textwidth}
        \centering
        \includegraphics[width=1.00\linewidth]{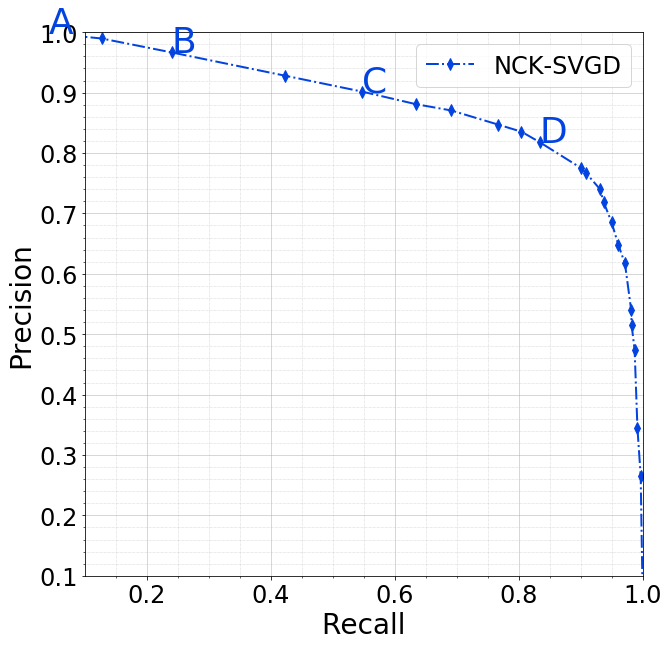}
        \caption{IPR of different $\beta$}
        \label{fig:mnist-beta-ipr}
    \end{subfigure}
    \begin{subfigure}{0.47\textwidth}
        \centering
        \includegraphics[width=1.0\linewidth]{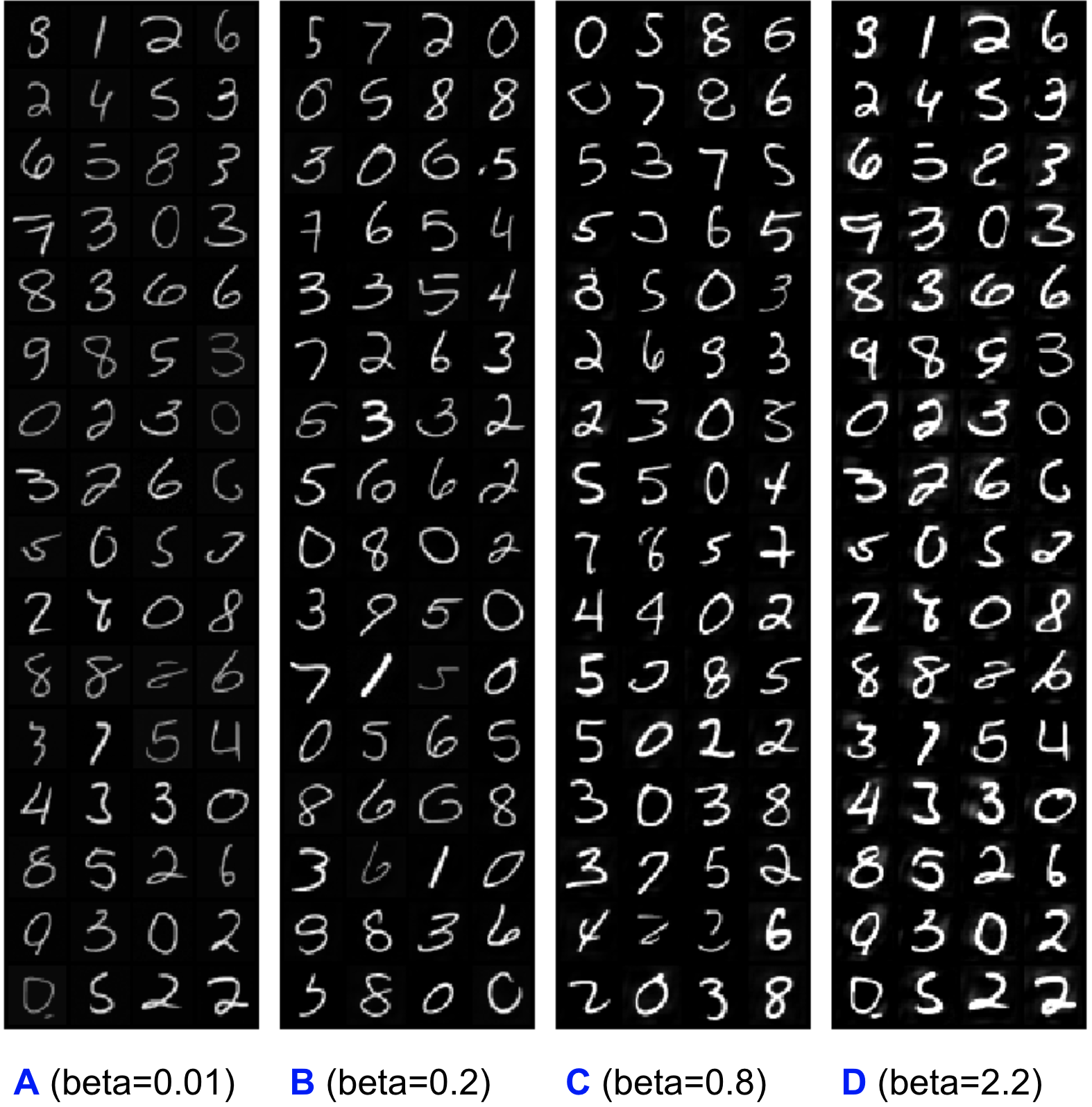}
        \caption{Corrsponding samples of different $\beta$}
        \label{fig:mnist-beta-samples}
    \end{subfigure}
    \caption{Varying entropy regularizer $\beta$ on the Improved Precision Reccall (IPR) curve and its corresponding samples.
    For \textcolor{blue}{A}, the $\beta=0.01$ and $(P,R)=(0.99, 0.04)$.
    For \textcolor{blue}{B}, the $\beta=0.2$ and $(P,R)=(0.96, 0.24)$.
    For \textcolor{blue}{C}, the $\beta=0.8$ and $(P,R)=(0.90, 0.54)$.
    For \textcolor{blue}{D}, the $\beta=2.2$ and $(P,R)=(0.81, 0.83)$.
    The empirical observation on MNIST's IPR curve aligns well with our theoretical insights on the entropy-regularization $\beta$.
    }
    \label{fig:mnist-beta}
    \vspace{-.5em}
\end{figure*}
From the MNIST's IPR curve in Figure~\ref{fig:mnist-beta}, We see the empirical evidence that supports the theoretical insights on the entropy-regularizer $\beta$ of \NCKSVGD.
We visualize four representative samples on the IPR curve, namely $A, B, C, D$, corresponding to different $\beta$ of $0.01, 0.2, 0.8, 2.2$, respectively. 
When $\beta$ is small (e.g., point \textcolor{blue}{A}), the precision is high and recall is small. The resulting samples show that many modes are disappearing.
In contrary, when $\beta$ is large (e.g., point \textcolor{blue}{D}), the precision becomes lower but recall is greatly increase. 
The resulting samples have better coverage in different digits.

\subsection{The Effect of Entropy Regularization on 2D Mixture of Gaussian}
\label{apx:entropy-reg-gmm}

\begin{figure*}
    \centering
    \begin{subfigure}{0.24\textwidth}
        \centering
        \includegraphics[width=1.00\linewidth]{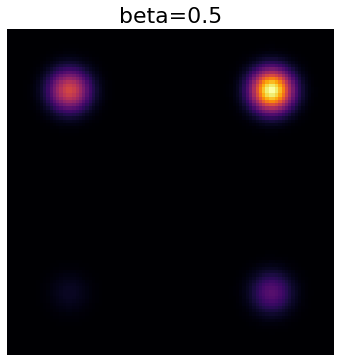}
        \caption{$\beta=0.5$}
        \label{fig:2d-gmm-beta-v0}
    \end{subfigure}
    \begin{subfigure}{0.24\textwidth}
        \centering
        \includegraphics[width=1.00\linewidth]{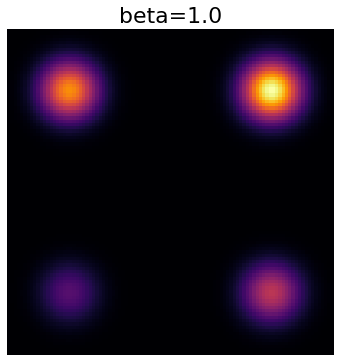}
        \caption{$\beta=1.0$}
        \label{fig:2d-gmm-beta-v1}
    \end{subfigure}
    \begin{subfigure}{0.24\textwidth}
        \centering
        \includegraphics[width=1.00\linewidth]{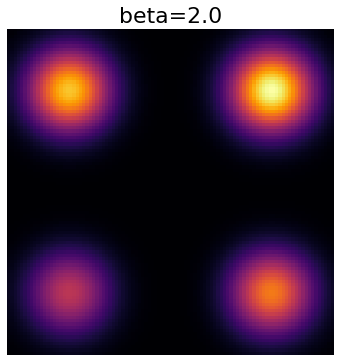}
        \caption{$\beta=2.0$}
        \label{fig:2d-gmm-beta-v2}
    \end{subfigure}
    \begin{subfigure}{0.24\textwidth}
        \centering
        \includegraphics[width=1.00\linewidth]{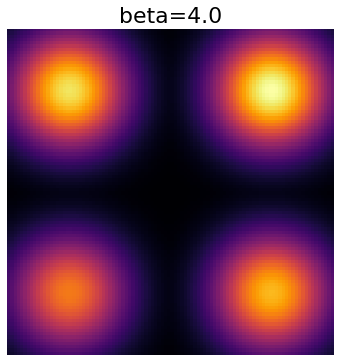}
        \caption{$\beta=4.0$}
        \label{fig:2d-gmm-beta-v3}
    \end{subfigure}
    \caption{(non-normalized) density of $p^{1/\beta}$ where $p(\xb)$ is a 2-dimensional mixture of Gaussian with imbalance mixture weights.}
    \vspace{-.5em}
    \label{fig:2d-gmm-beta}
\end{figure*}
As shown in Figure~\ref{fig:2d-gmm-beta}, we visualize the non-normalized density of a 2-dimensional Gaussian mixture $p(\xb)^{1/\beta}$ with varying choices of the entropy regularizer $\beta$.
Specifically, consider $p(\xb) = 0.8\Ncal\big((5,5), I\big) + 0.2\Ncal\big((-5,-5), I\big) + 0.6\Ncal\big((5,-5), I\big) + 0.4\Ncal\big((-5,5), I\big)$.
When $\beta$ is small (e.g., $\beta=0.5$), the resulting $p^{1/\beta}$ shows mode dropping compared to original $p$.
When $\beta$ is large (e.g., $\beta=4.0$), the resulting $p^{1/\beta}$ covers all four modes, but wrongly  with the almost equal weights.

\newpage
\section{Additional Experiment Details}

\subsection{Toy Experiment}
\label{app:exp-setup-toy}
For the results in Figure~\ref{fig:toy-gmm}, we mainly follow the setting of~\cite{song2019generative} with $p_d(\xb) = 0.2\Ncal((-5,-5), I)+0.8\Ncal((5,5), I)$.
We generate $1024$ samples for each subfigure of Figure~\ref{fig:toy-gmm}.
The score function can be analytically derived from $p_d$.
The initial samples are all uniformly chosen
in the square $8 \times 8$.
For \SGLD and \SVGD, we use $T = 1000$.
For \ASGLD, \ASVGD, \NCKSVGD, we use $T = 100$, $L = 10$, $\sigma_1=20.0$, $\sigma_{10}=1.0$.
The learning rate $\epsilon$ is chosen from $\{0.1, 0.5, 1.0, 2.0, 4.0, 8.0, 16.0\}$.
When evaluating with Maxmimum Mean Discrepancy $\MMD_k(P_d, Q)$ between the real data samples from $p_d$ and the generated samples from different sampling methods $Q$,
we consider the RBF kernel and set bandwidth by median heuristic.
The experiment was run on one Nvidia 2080Ti GPU.

\subsection{Image Generation}
\label{app:exp-setup-real}

\paragraph{Network Architecture}
For the noise-conditional score network, we use the pre-trained model~\cite{song2019generative}\footnote{\url{https://github.com/ermongroup/ncsn}}. 
For the noise-conditional kernel, we consider a modified NCSN architecture where the encoder consists of ResNet with instance normalization layer, and the decoder consists of U-Net-type architecture. 
The critical difference to the score network is the dimension of bottleneck layer $h$, which is $h=196$ for MNIST and $j=512$ for CIFAR-10.
Note that $h$ for both MNIST and CIFAR-10 are considerably smaller than the data dimension, which is $d=768$ for MNIST and $d=3072$ for CIFAR-10.
In contrast, the dimension of the hidden layers of NCSN is around $4$x larger than the data dimension. 

\paragraph{Kernel Design}
We consider a Mixture of RBF and IMQ kernel on the data-space and code-space features, as defined in Eq~\eqref{eq:deep-mok}.
The bandwidth of RBF kernel $\gamma(\sigma) = \gamma_0 / med(X_\sigma)$,
where $med(X_\sigma)$ denotes the median of samples' pairwise distance drawn from anneal data distributions $p_\sigma(\tilde{\xb}|\xb)$.
We search for the best kernel hyper-parameters $\gamma_0=\{0.4, 0.6, 0.8, 1.0, 2.0, 4.0\}$ and $\tau_0=\{-0.1, -0.2, -0.3, -0.4\}$. 

\paragraph{Inference Hyper-parameters}
Following \cite{song2019generative}, we choose $L=10$ different noise levels where the standard deviations $\{\sigma_i\}_{i=1}^L$ is a geometric sequence with $\sigma_1=1$ and $\sigma_{10}=0.01$.
Note that Gaussian noise of $\sigma=0.01$ is almost indistinguishable to human eyes for image data.
For \ASGLD, we choose $T=100$ and $\epsilon=2\times10^{-5}$ and $\alpha=\{0.1, 0.2, \ldots, 1.1, 1.2\}$.
For \NCKSVGD, we choose $n=128$, $T=50$, $\epsilon=\{2,4,6\} \times10^{-4}$, $\beta=\{0.01, 0.05, 0.1, 0.2, \ldots, 1.0, \ldots, 3.9, 4.0\}$.

\paragraph{Evaluation Metric}
We report the
Inception~\cite{salimans2016improved}\footnote{\url{https://github.com/openai/improved-gan/tree/master/inception_score}}
and FID~\cite{heusel2017gans}\footnote{\url{https://github.com/bioinf-jku/TTUR}}
scores using $50$k samples.
In addition, We also present the Improved Precision Recall (IPR) curve~\cite{kynkaanniemi2019improved}\footnote{\url{https://github.com/kynkaat/improved-precision-and-recall-metric}}
to justify the impact of entropy regularization and kernel hyper-parameters on diversity versus quality trade-off.
For the IPR curve on MNIST, we use data-space features (i.e., raw image pixels) to compute the KNN-3 data manifold, as gray-scale images do not apply to the VGG-16 network.
For the IPR curve on CIFAR-10, we follow the origin setting of~\cite{kynkaanniemi2019improved} that uses code-space embeddings from the pre-trained VGG-16 model to construct the KNN-3 data manifold.
For simplicity, we generate 1024 samples to compute the precision and recall, and report the average of 5 runs with different random seeds.

\newpage
\subsection{Baseline Comparison with \SVGD and \ASVGD}
\label{app:exp-baseline}

\begin{figure*}[!ht]
    \centering
    \begin{subfigure}{0.32\textwidth}
        \centering
        \includegraphics[width=1.00\linewidth]{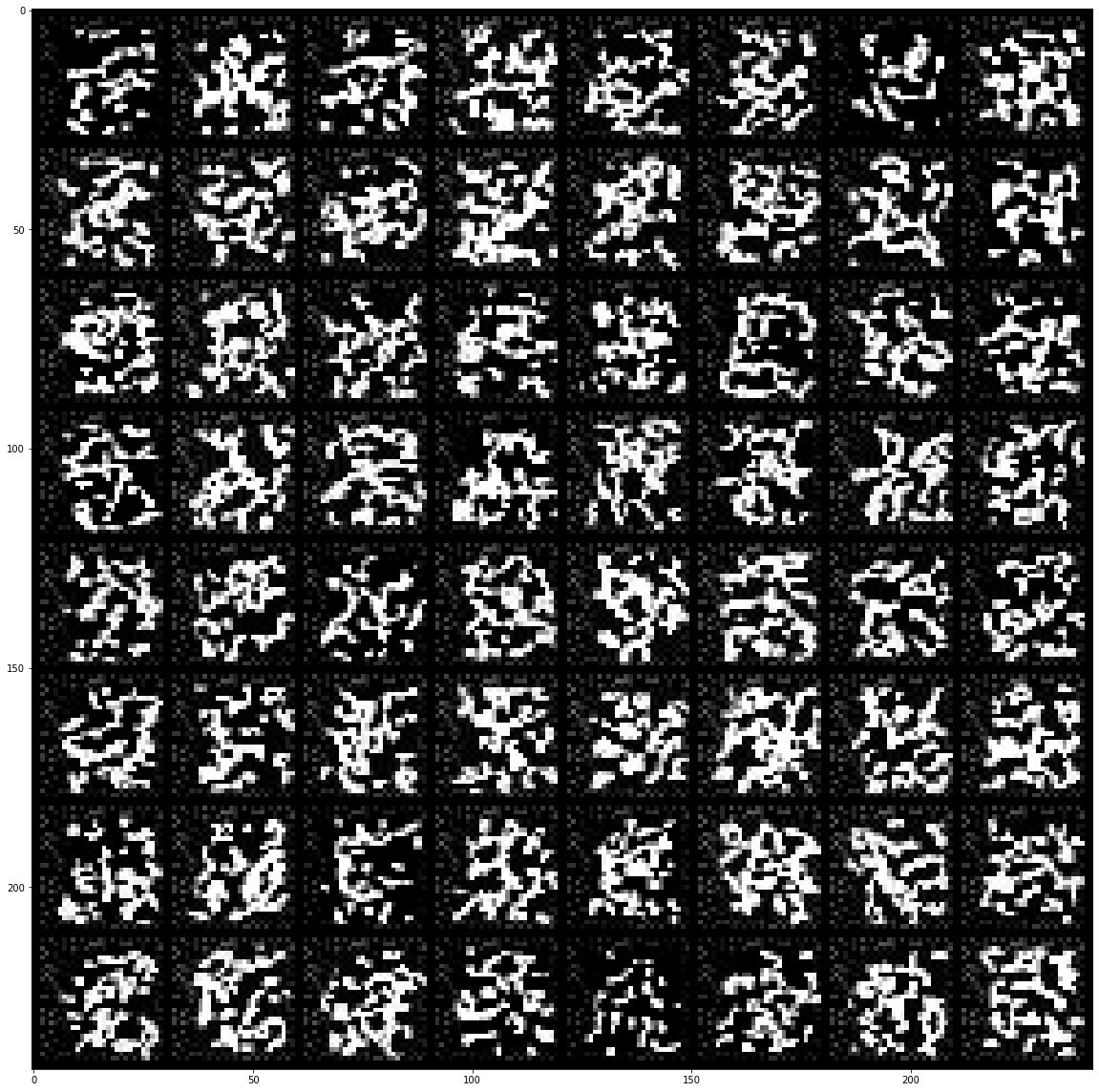}
        \caption{\SVGD}
        \label{fig:mnist_svgd-v0}
    \end{subfigure}
    \begin{subfigure}{0.32\textwidth}
        \centering
        \includegraphics[width=1.00\linewidth]{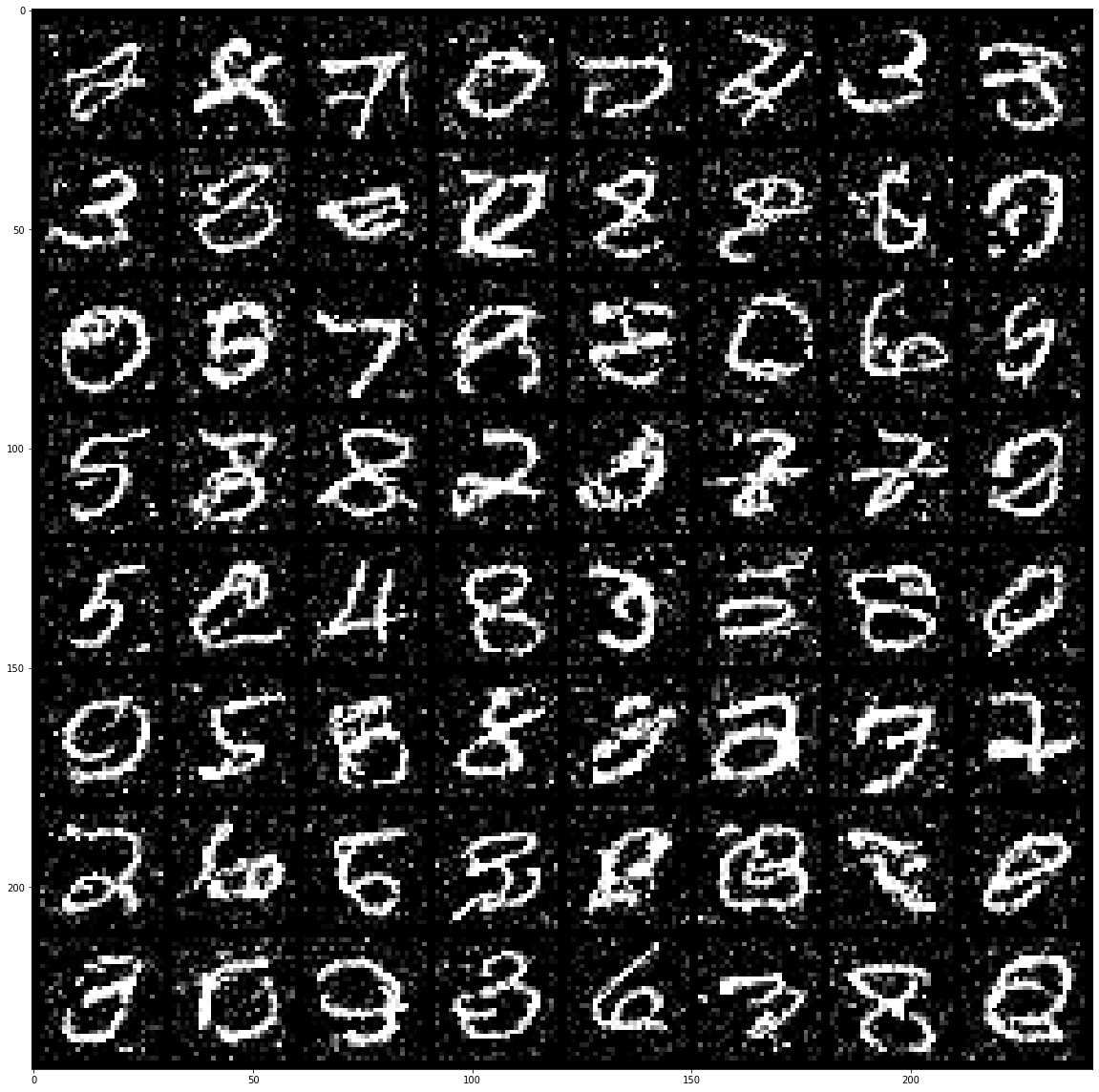}
        \caption{\ASVGD}
        \label{fig:mnist_svgd-v1}
    \end{subfigure}
    \begin{subfigure}{0.32\textwidth}
        \centering
        \includegraphics[width=1.00\linewidth]{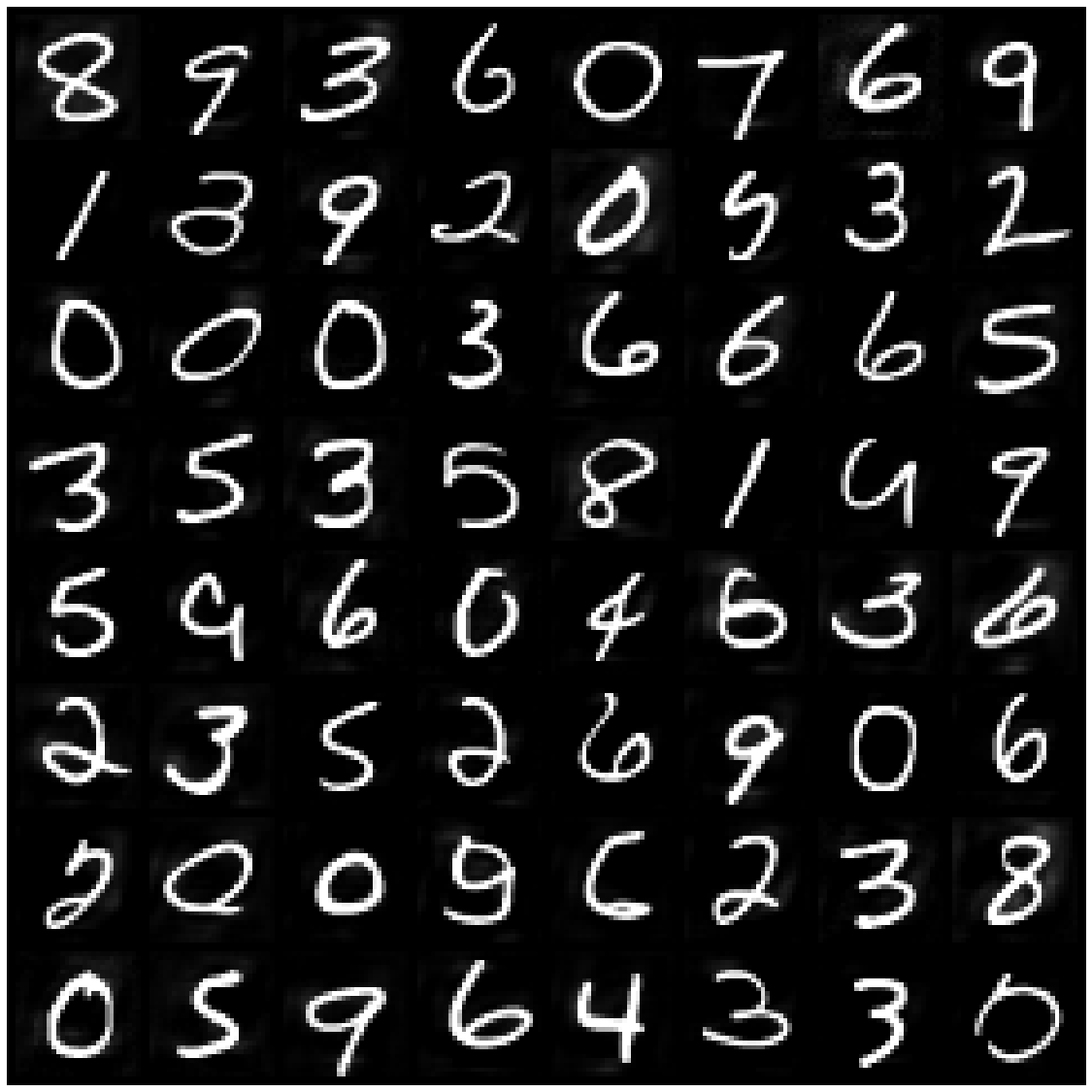}
        \caption{\NCKSVGD}
        \label{fig:mnist_svgd-v2}
    \end{subfigure}
    \caption{SVGD baseline comparison on MNIST.}
    \vspace{-1em}
    \label{fig:mnist-baseline-exp}
\end{figure*}
\begin{figure*}[!ht]
    \centering
    \begin{subfigure}{0.32\textwidth}
        \centering
        \includegraphics[width=1.00\linewidth]{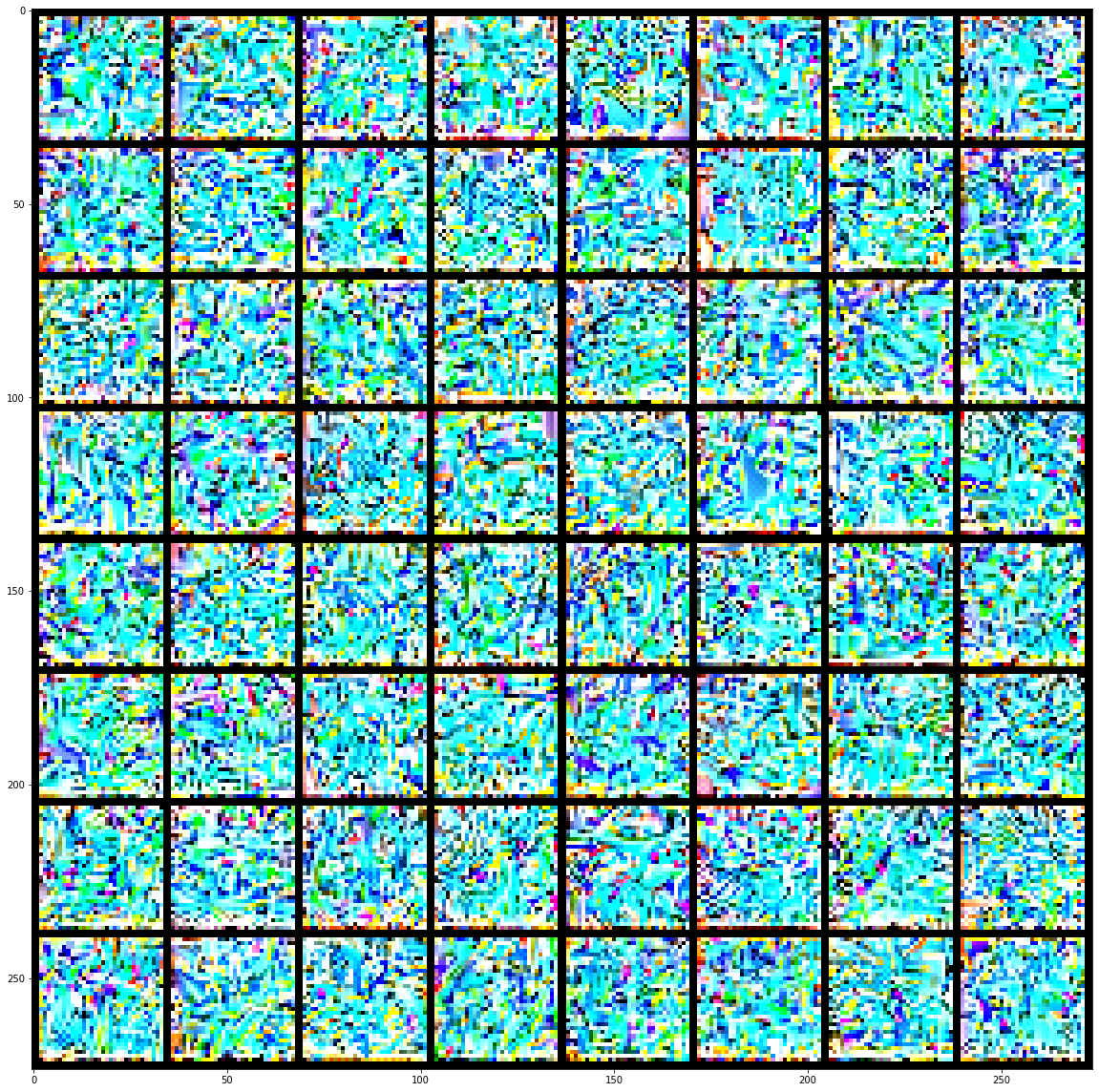}
        \caption{\SVGD}
        \label{fig:cifar10_svgd-v0}
    \end{subfigure}
    \begin{subfigure}{0.32\textwidth}
        \centering
        \includegraphics[width=1.00\linewidth]{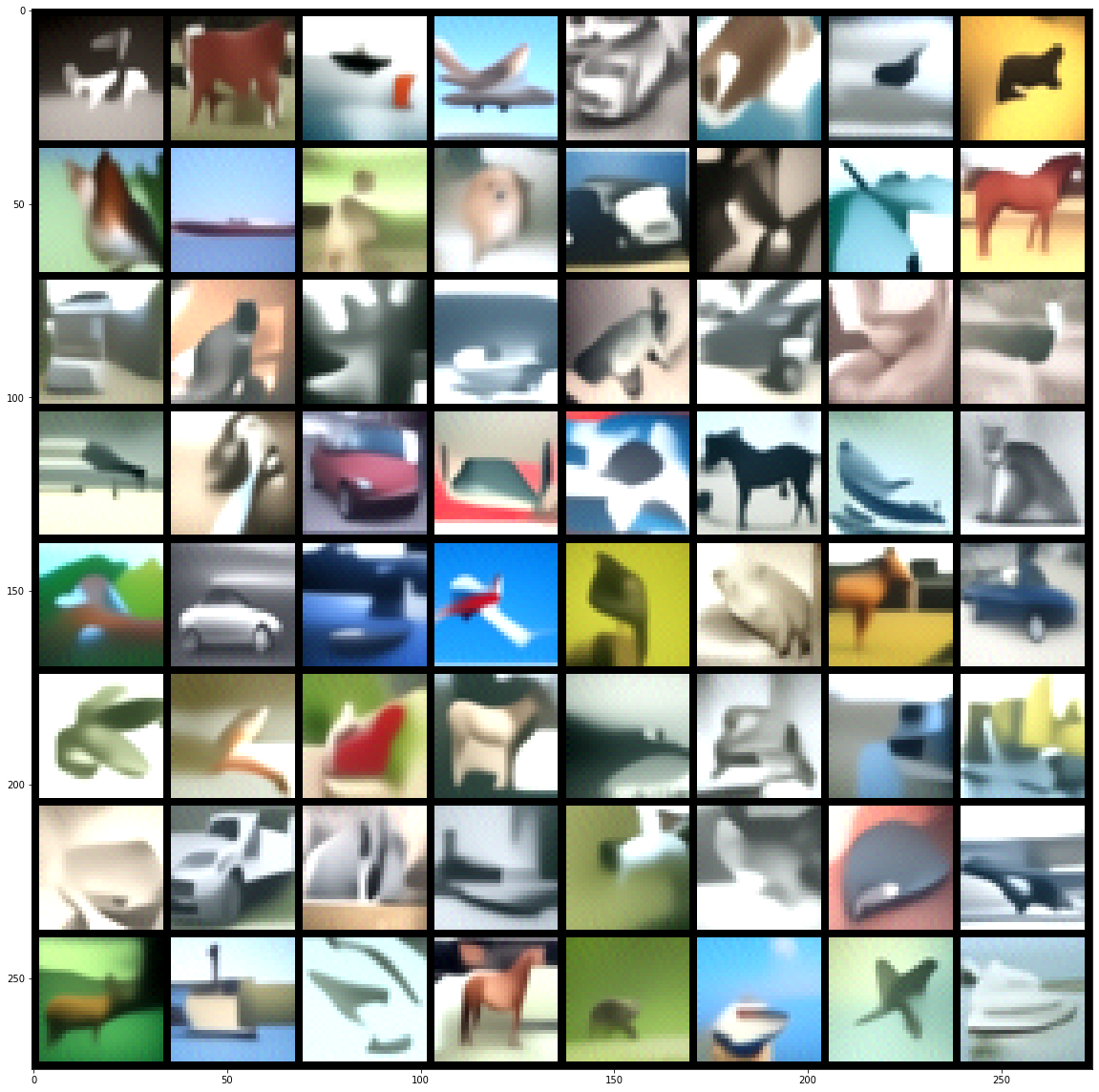}
        \caption{\ASVGD}
        \label{fig:cifar10_svgd-v1}
    \end{subfigure}
    \begin{subfigure}{0.32\textwidth}
        \centering
        \includegraphics[width=1.00\linewidth]{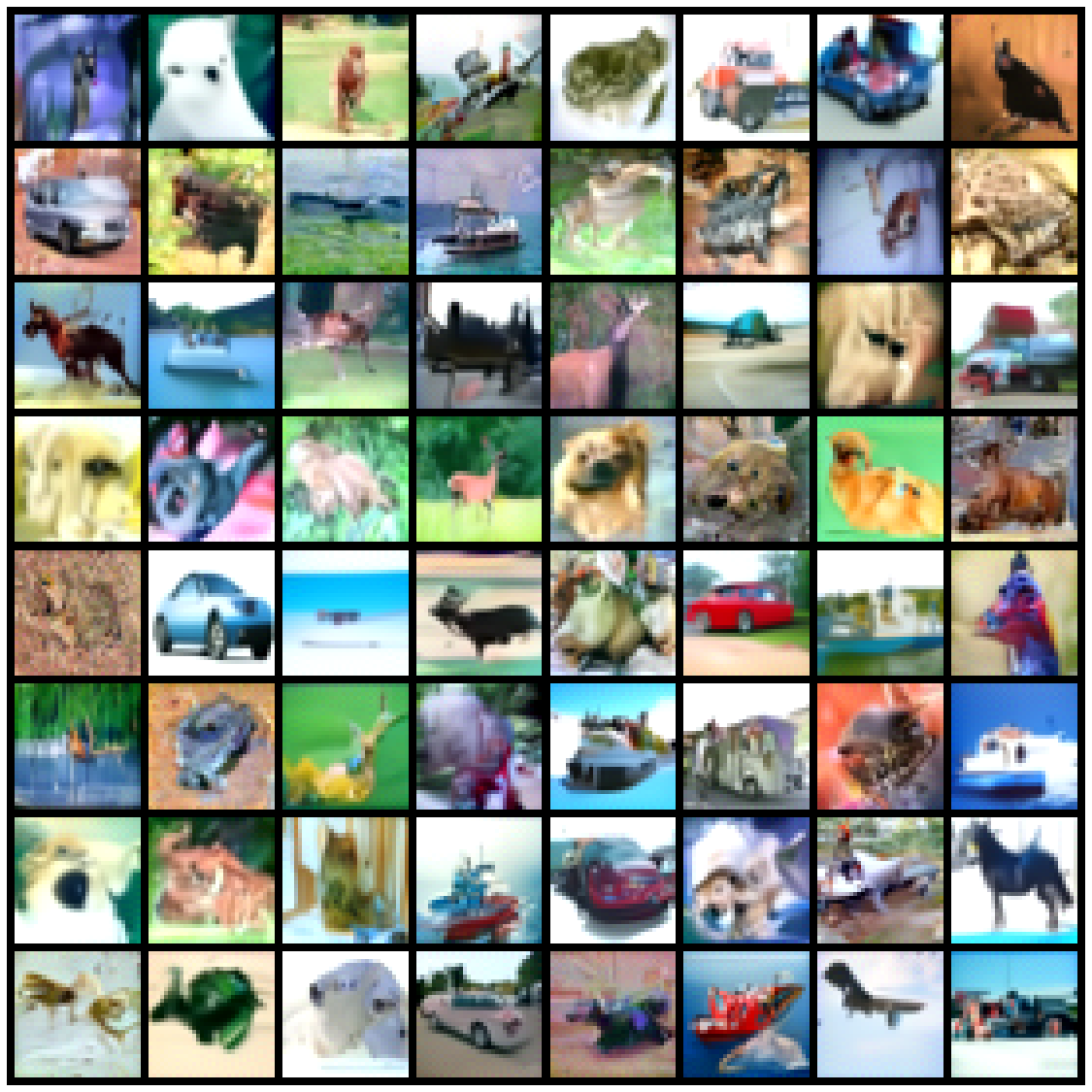}
        \caption{\NCKSVGD}
        \label{fig:cifar10_svgd-v2}
    \end{subfigure}
    \caption{SVGD baseline comparison on CIFAR-10.}
    \vspace{-1em}
    \label{fig:cifar10-baseline-exp}
\end{figure*}
\begin{figure*}[!ht]
    \centering
    \begin{subfigure}{0.32\textwidth}
        \centering
        \includegraphics[width=1.00\linewidth]{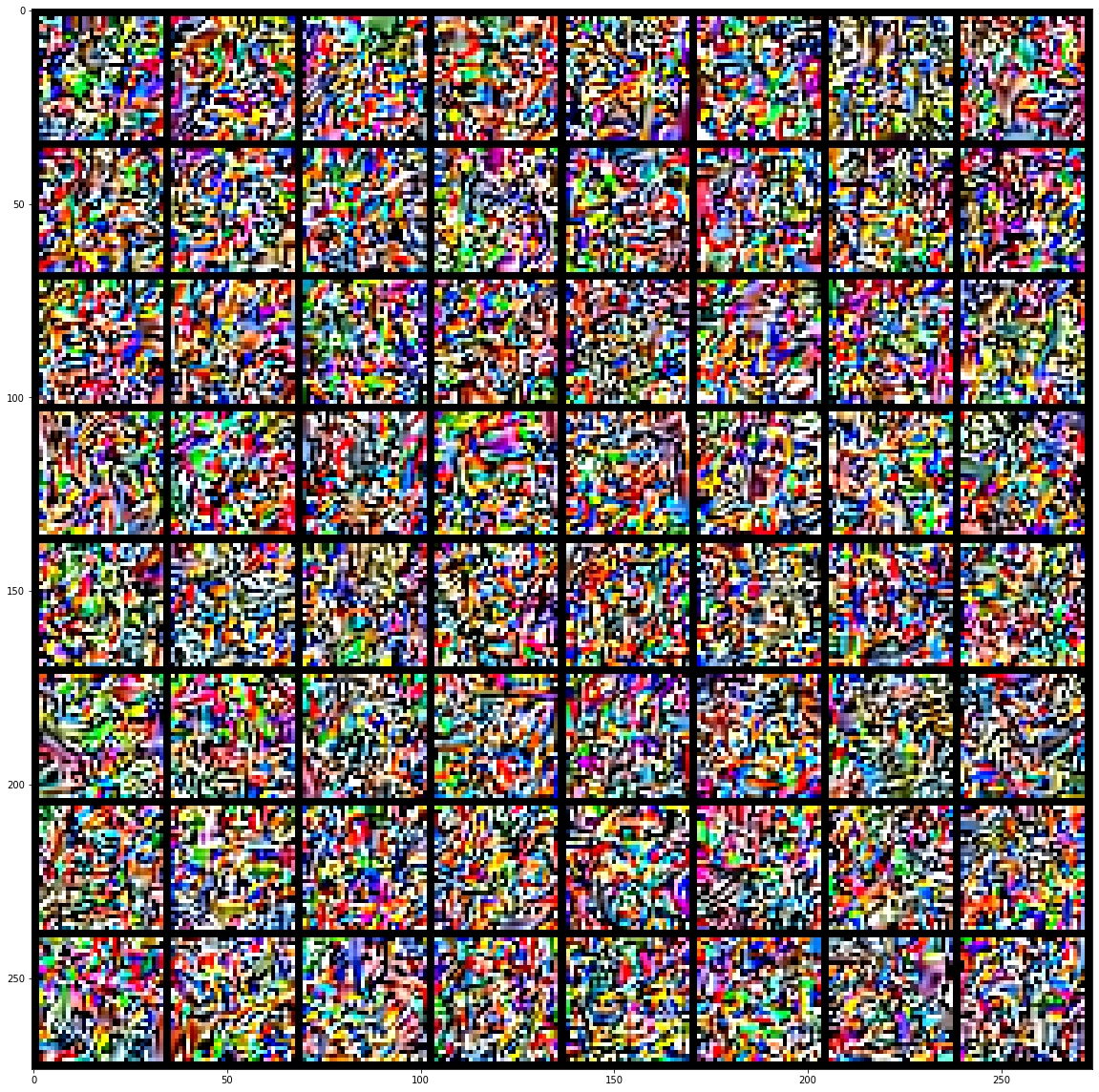}
        \caption{\SVGD}
        \label{fig:celeba_svgd-v0}
    \end{subfigure}
    \begin{subfigure}{0.32\textwidth}
        \centering
        \includegraphics[width=1.00\linewidth]{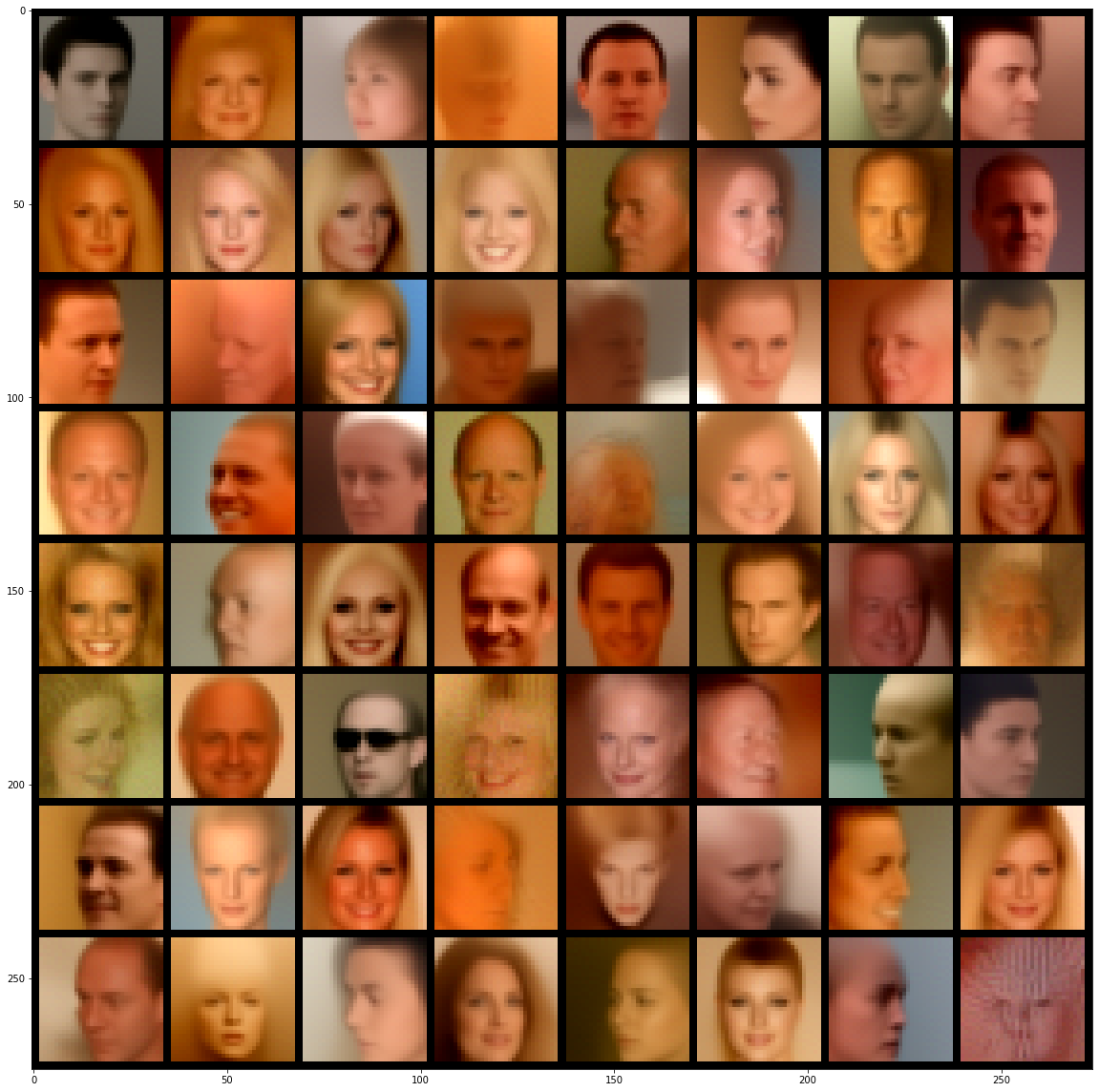}
        \caption{\ASVGD}
        \label{fig:celeba_svgd-v1}
    \end{subfigure}
    \begin{subfigure}{0.32\textwidth}
        \centering
        \includegraphics[width=1.00\linewidth]{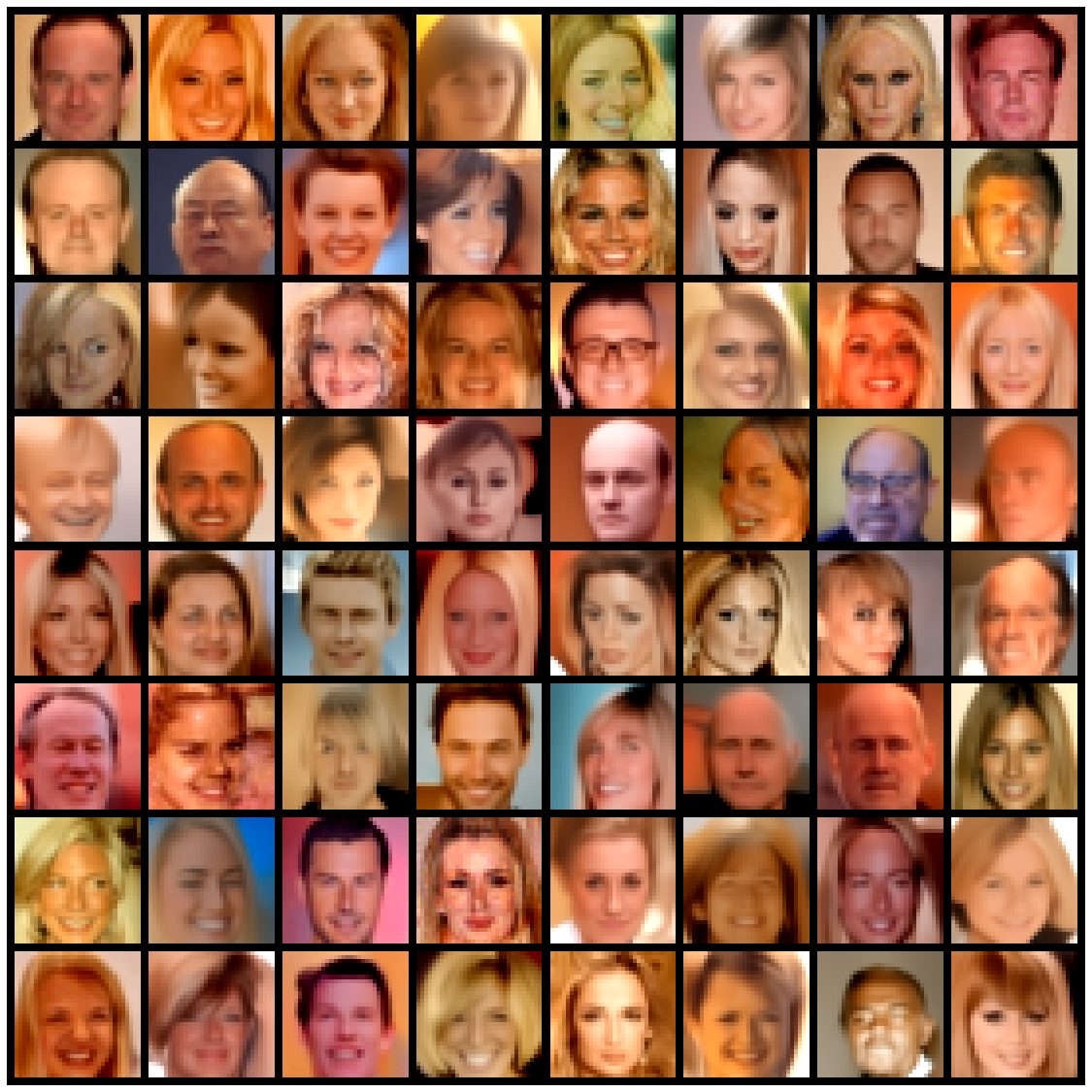}
        \caption{\NCKSVGD}
        \label{fig:celeba_svgd-v2}
    \end{subfigure}
    \caption{SVGD baseline comparison on CelebA.}
    \label{fig:celeba-baseline-exp}
\end{figure*}

Similar to the study in Section~\ref{sec:challenges}, we compare the proposed \NCKSVGD with two SVGD baselines, namely the vanilla SVGD (i.e., \SVGD) and anneal SVGD with a fixed kernel (i.e., \ASVGD), on three image generation benchmarks.
See MNIST results in Figure~\ref{fig:mnist-baseline-exp}, and CIFAR-10 results in Figure~\ref{fig:cifar10-baseline-exp}, and CelebA results in Figure~\ref{fig:celeba-baseline-exp}.
We present the qualitative study only and omit the quantitative evaluation, as the performance difference can be clearly distinguish from the sample quality alone.
We can see that the proposed \NCKSVGD produces higher quality samples comparing against two baselines, \SVGD and \ASGLD.

\end{document}